\documentclass[letterpaper]{article} % DO NOT CHANGE THIS
\usepackage{aaai2026}  % DO NOT CHANGE THIS
\usepackage{times}  % DO NOT CHANGE THIS
\usepackage{helvet}  % DO NOT CHANGE THIS
\usepackage{courier}  % DO NOT CHANGE THIS
\usepackage[hyphens]{url}  % DO NOT CHANGE THIS
\usepackage{graphicx} % DO NOT CHANGE THIS
\usepackage{natbib}  % DO NOT CHANGE THIS AND DO NOT ADD ANY OPTIONS TO IT
\usepackage{caption} % DO NOT CHANGE THIS AND DO NOT ADD ANY OPTIONS TO IT
\usepackage{algorithm}
\usepackage{algorithmic}
\usepackage{subcaption}
\usepackage{newfloat}
\usepackage{listings}
\DeclareCaptionStyle{ruled}{labelfont=normalfont,labelsep=colon,strut=off} % DO NOT CHANGE THIS
\floatstyle{ruled}
\newfloat{listing}{tb}{lst}{}
\floatname{listing}{Listing}
\usepackage{amsfonts}
\usepackage{xspace}
\usepackage{amsmath}
\usepackage{multirow}
\usepackage{booktabs}
\newcommand{\eg}{\emph{e.g.}\xspace}
\newcommand{\ie}{\emph{i.e.}\xspace}
\newcommand{\vs}{\emph{vs.}\xspace}

\title{Difficulty-Aware Label-Guided Denoising for Monocular 3D Object Detection}
\author{
    Soyul Lee\textsuperscript{\rm 1}\equalcontrib $\quad$
    Seungmin Baek\textsuperscript{\rm 1}\equalcontrib $\quad$
    Dongbo Min\textsuperscript{\rm 1}\thanks{Corresponding author.}
}
\affiliations{
    \textsuperscript{\rm 1}School of AI and Software, Ewha Womans University, South Korea\\
    \{230aig, min100kim, dbmin\}@ewha.ac.kr

}

\begin{document}

\maketitle

\begin{abstract}
Monocular 3D object detection is a cost-effective solution for applications like autonomous driving and robotics, but remains fundamentally ill-posed due to inherently ambiguous depth cues. Recent DETR-based methods attempt to mitigate this through global attention and auxiliary depth prediction, yet they still struggle with inaccurate depth estimates. Moreover, these methods often overlook instance-level detection difficulty, such as occlusion, distance, and truncation, leading to suboptimal detection performance. We propose MonoDLGD, a novel \textbf{D}ifficulty-Aware \textbf{L}abel-\textbf{G}uided \textbf{D}enoising framework that adaptively perturbs and reconstructs ground-truth labels based on detection uncertainty. Specifically, MonoDLGD applies stronger perturbations to easier instances and weaker ones into harder cases, and then reconstructs them to effectively provide explicit geometric supervision. By jointly optimizing label reconstruction and 3D object detection, MonoDLGD encourages geometry-aware representation learning and improves robustness to varying levels of object complexity. Extensive experiments on the KITTI benchmark demonstrate that MonoDLGD achieves state-of-the-art performance across all difficulty levels.
\end{abstract}
\begin{links}
    \link{Code}{https://github.com/lsy010857/MonoDLGD}
    % \link{Datasets}{https://aaai.org/example/datasets}
    % \link{Extended version}{https://aaai.org/example/extended-version}
\end{links}

%%%%%%%%%%%%%%%%%%%%%%%%%%%%%%%%%%%%%%%%%%%%%%%%%%

\section{Introduction}

Monocular 3D object detection aims to estimate the 3D location, size, and orientation of objects using a single RGB image. Owing to its low cost, ease of deployment, and compatibility with high-resolution images, it has become an attractive solution for applications such as autonomous driving, robotics, and augmented reality. However, unlike LiDAR-based~\cite{lion,dsvt,PointPillar} or stereo-based~\cite{dsgn++,liga,Stereo_R-CNN} methods, monocular approaches inherently suffer from a lack of depth cues, rendering 3D geometry estimation fundamentally ill-posed.

\begin{figure}[htb]
    \centering
    \includegraphics[width=0.9\linewidth]{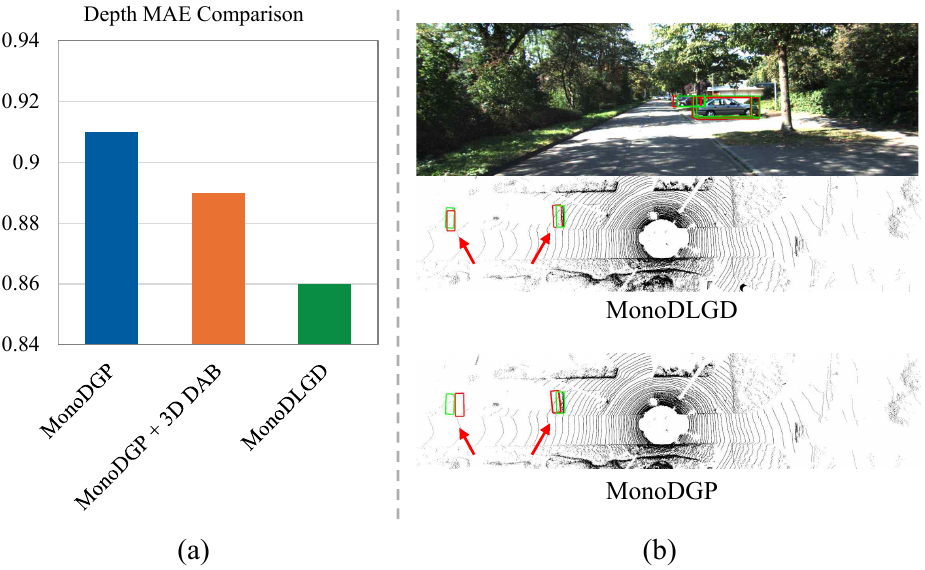}
    \caption{ \textbf{Depth-Centric Detection: Elevating Depth and BEV Accuracy via MonoDLGD.} 
    {(a)} Depth estimation accuracy using mean absolute error (MAE) on the KITTI validation set for MonoDGP~\cite{monodgp}, MonoDGP with our 3D-DAB, and ours. {(b)} Bird's-eye view (BEV) visualization. MonoDLGD achieves more accurate and robust detection across varying object distances, highlighting its improved geometric understanding.}
    \label{fig:depth_mae_comparison}
\end{figure}

Recent advances in monocular 3D object detection have been driven by the adaptation of Transformer-based architectures, particularly the DEtection TRansformer (DETR)~\cite{detr}, originally developed for 2D object detection. MonoDETR~\cite{monodetr} first introduces DETR into monocular 3D detection, moving beyond CenterNet-based approaches~\cite{monodle,gpenet,monoflex} that focus on local features. By leveraging the global attention mechanism of Transformers, it effectively captures spatial and depth relationships between objects. To compensate for the absence of explicit depth cues, MonoDETR~\cite{monodetr} and MonoDGP~\cite{monodgp} incorporate auxiliary depth prediction heads, injecting geometric priors into the detection pipeline. However, since the depth estimation relies solely on a single image, these methods remain constrained by the ill-posed nature of monocular images, which limits the accuracy of depth prediction. As illustrated in Figure~\ref{fig:depth_mae_comparison}, this inherent limitation results in considerable errors in 3D object localization.

{MonoMAE~\cite{monomae} attempt to improve robustness to occlusion by masking and reconstructing object features based on their depth levels, thereby enabling better 3D representation learning for partially visible objects. While effective to some extent, its difficulty modeling is limited to isolated factors such as occlusion status or depth range. In monocular settings, however, detection difficulty arises from a combination of factors including object scale, distance, truncation, and occlusion. Ignoring this multi-factor complexity can degrade both training stability and representation quality~\cite{monoflex,monomae}.}

{To address the fundamental limitations of monocular 3D detection, we propose Difficulty-Aware Label-Guided Denoising (MonoDLGD), a novel framework that injects adaptive perturbations into ground-truth labels and learns to reconstruct them, providing explicit geometric supervision during training. Unlike prior DETR-based methods~\cite{monodetr,monodgp,monomae}, our approach directly operates on 3D ground truth labels, enabling more stable and geometry-aware representation learning. MonoDLGD introduces two key components: (1) a denoising strategy that perturbs and reconstructs ground-truth labels containing rich 3D information, and (2) a difficulty-aware perturbation (DAP) mechanism that modulates the strength of perturbations based on instance-level detection difficulty. Together, these components guide the model to learn robust geometric representations across objects with diverse complexities, with only a marginal inference-time overhead.}

\begin{figure}[t!]
    \centering
    \includegraphics[width=0.9\linewidth]{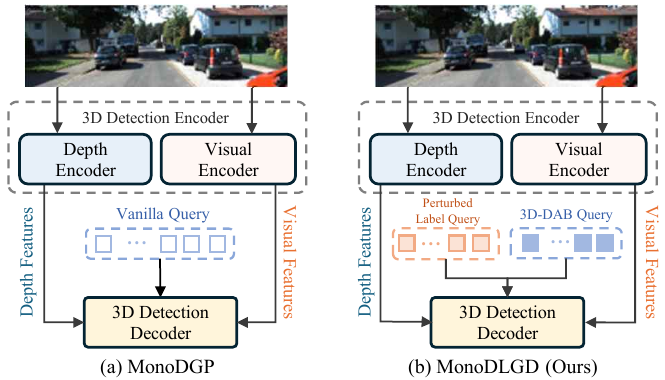}
    \caption{\textbf{Structural Comparison with MonoDGP~\cite{monodgp}.} Our method introduces 3D-DAB queries to encode spatial priors and explicitly provides 3D geometric supervision by reconstructing perturbed label queries within a shared decoder.}
    \label{fig:arich_comparison}
\end{figure}

{Specifically, MonoDLGD perturb ground-truth labels such as projected bounding boxes and depths during training and learns to reconstruct them via a shared decoder, as illustrated in Fig.~\ref{fig:arich_comparison} (b). This denoising process provides strong supervision signals that help the model better understand 3D structure from monocular cues. To further enhance this effect, we introduce 3D Dynamic Anchor Box (3D-DAB), which embeds spatial priors (object projections and depths) into the queries, tightly aligning it with the perturbed label representations in the decoder. The reconstruction of these perturbed labels enables the decoder to transfer geometric signals into the detection pipeline more effectively.}

{Crucially, not all objects are equally difficult to detect in monocular settings. Small, distant, or occluded instances present greater ambiguity, and applying uniform perturbations may degrade their structural signals. To address this, MonoDLGD estimates instance-wise uncertainty as a proxy for detection difficulty and adaptively scales perturbation strength. Hard instances receive weaker perturbations to preserve their geometry, while easier ones are perturbed more aggressively. This difficulty-aware strategy promotes stable and discriminative feature learning across varying levels of complexity, ultimately improving 3D detection accuracy.}

Our main contributions are as follows:
\begin{itemize}

    \item We propose \textbf{Difficulty-Aware Label-Guided Denoising (MonoDLGD)}, which introduces label perturbation and reconstruction guided by prediction uncertainty, effectively leveraging explicit geometric supervision. %to address the ill-posed nature of monocular 3D detection.

    \item We show that \textbf{modeling instance-level uncertainty} alone substantially improves detection accuracy, highlighting the importance of uncertainty-aware denoising in monocular 3D object detection.
        
    \item {Our method achieves \textbf{state-of-the-art performance} on the KITTI benchmark without any additional inference overhead, as the difficulty-aware pertubation and reconstruction are confined to the training phase.}

\end{itemize}

%%%%%%%%%%%%%%%%%%%%%%%%%%%%%%%%%%%%%%%%%%

\section{Related Work}

% subfigure -> minipage
\begin{figure*}
  \centering
  \begin{minipage}[t]{0.62\linewidth}
    \centering
    \includegraphics[width=\linewidth]{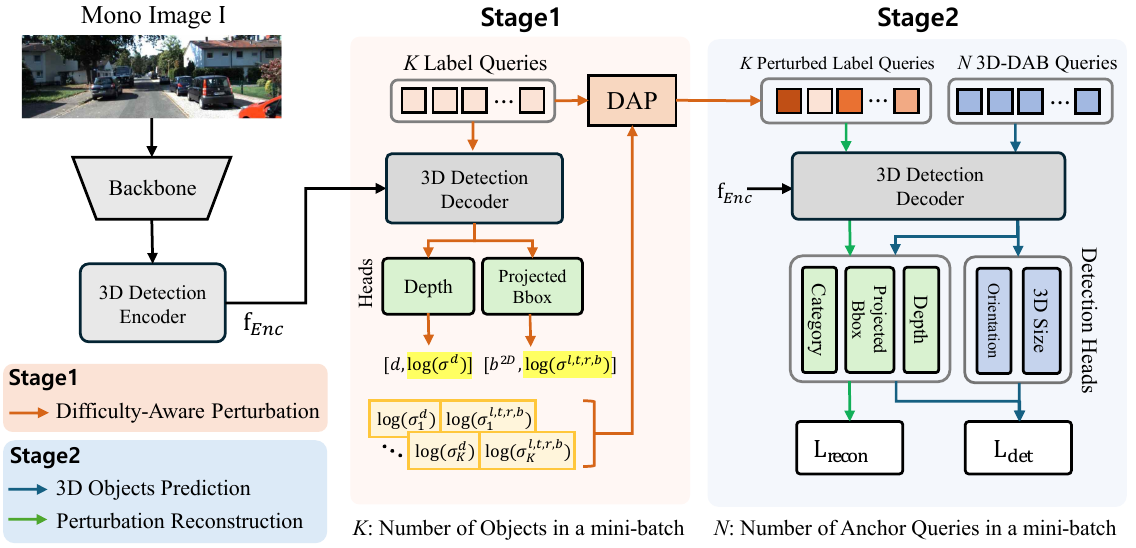}    
    \par\small (a) Overall architecture of MonoDLGD
    \label{fig:MonoDLGD_arch}
  \end{minipage} 
  \hfill
  \begin{minipage}[t]{0.24\linewidth}
    \centering
    \includegraphics[width=\linewidth]{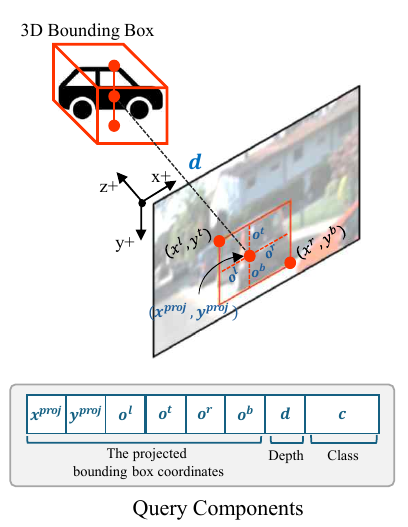} 
    \par\small (b) Query Component Structure
  \end{minipage}
  \caption{Overview of the proposed MonoDLGD: (a) MonoDLGD adopts a two-stage architecture after extracting the encoder feature $f_{Enc}$ containing depth and visual features. %from the 3D detection encoder.
  \textbf{Stage 1} (red arrows) performs Difficulty-Aware Perturbation (DAP) by first estimating the uncertainty of bounding box ($\sigma^{l}$, $\sigma^{t}$, $\sigma^{r}$, $\sigma^{b}$) and depth ($\sigma^d$) attributes and then adaptively perturbing the label queries based on the estimated uncertainties. \textbf{Stage 2} (blue and green arrows) feeds both the perturbed label queries and the 3D-DAB queries into the decoder. (b) illustrates the internal components of label queries and 3D-DAB queries, all of which share a common structure.}
  \label{fig:MonoDLGD_overview}
\end{figure*}

\subsection{Monocular 3D Detection with Transformers}
Recent advances in monocular 3D object detection have been largely driven by the adoption of Transformer architectures, owing to their superior ability to capture global context and long-range dependencies compared to CNN-based approaches~\cite{monodle,gpenet,monoflex}. MonoDETR~\cite{monodetr} introduced the first DETR-style monocular 3D detector with a depth-guided transformer, using a predicted depth map and depth-aware decoder to incorporate global context. MonoDGP~\cite{monodgp} further improved transformer detectors by decoupling 2D and 3D query streams and introducing perspective-invariant geometry error priors to refine depth estimation. Meanwhile, MonoMAE~\cite{monomae} tackled occlusion via a depth-aware masked autoencoder: it masks parts of object queries and learns to reconstruct them, thereby handling heavily occluded objects. 
However, these methods inherently suffer from ill-posed geometric constraints, as depth estimation relies solely on a single image. Moreover, they tend to overlook instance-level challenges such as occlusion, distance, and truncation, leading to unstable training and degraded 3D representation quality.

\subsection{Denoising Strategies for Object Detection}
Denoising has recently emerged as an effective technique to stabilize training and enhance detection performance, especially within Transformer-based detection frameworks. DN-DETR~\cite{dndetr} introduced denoising to DETR by perturbing ground-truth boxes and training the model to reconstruct them, significantly accelerating training convergence and reducing instability during bipartite matching. This approach has been further improved by DINO~\cite{zhang2022dino}, which introduced contrastive denoising by leveraging noisy queries to explicitly model positive and negative query pairs, further enhancing detection accuracy and training stability. Denoising techniques have also been extended to 3D detection. ConQueR~\cite{conquer} applies denoising to LiDAR-based detectors by perturbing and reconstructing queries in the voxel space to achieve sparse predictions and SEED~\cite{seed} adopts denoising training in point cloud-based DETR frameworks, enhancing detection accuracy and robustness for point cloud data.
In this paper, we leverage a denoising strategy to explicitly incorporate geometric supervision. Unlike previous methods that apply uniform denoising, our approach directly injects adaptive perturbations into ground-truth labels based on instance-level detection difficulty, enabling the model to stably learn robust 3D representations. % across varying levels of object complexity.

\subsection{Uncertainty Estimation}
\cite{kendall2017uncertainties} formally categorize uncertainty into aleatoric uncertainty, which captures noise inherent in observations, and epistemic uncertainty, which accounts for uncertainty in model parameters. The aleatoric uncertainty has been actively explored in the context of object detection~\cite{yolov3_gaussian,monopair,he2019bounding,monoflex}. For example, Gaussian YOLOv3~\cite{yolov3_gaussian} models bounding boxes as Gaussian distributions to quantify localization uncertainty, which helps to rectify detection scores and reduce false positives. ~\cite{he2019bounding} predict bounding boxes as Gaussian distributions and utilize KL divergence as the regression loss, explicitly accounting for aleatoric uncertainty to achieve accurate object detection. MonoPair~\cite{monopair} leverages uncertainty to weight pair-wise geometric constraints during post-processing optimization, effectively improving detection stability and accuracy. MonoFlex~\cite{monoflex} further models uncertainties of depth estimates from multiple depth predictors, leveraging aleatoric uncertainty to adaptively combine depth predictions from direct regression and keypoint-based geometry, significantly improving localization accuracy. 
In this paper, we leverage aleatoric uncertainty to enhance both training stability and representation quality in monocular 3D object detection. We incorporate this uncertainty into the denoising process for both 2D bounding box and depth supervision and use it to guide the adaptive perturbation strength in our difficulty-aware label-guided denoising framework.

%%%%%%%%%%%%%%%%%%%%%%%%%%%%%%%%%%%%%%%%%%%%%%%
%%                 Method                    %%  
%%%%%%%%%%%%%%%%%%%%%%%%%%%%%%%%%%%%%%%%%%%%%%%
\section{Method}

\subsection{Motivation and Overview} 

Existing DETR-based monocular 3D object detectors, such as  MonoDETR~\cite{monodetr} and MonoDGP~\cite{monodgp}, leverage auxiliary foreground depth maps to alleviate depth ambiguity but still fundamentally suffer from the ill-posed nature of monocular geometry. These methods also uniformly treat all objects during training, overlooking key difficulty factors such as object size, distance, occlusion, and truncation. MonoMAE~\cite{monomae} partially addresses these issues via occlusion-aware masking, but its consideration of detection difficulty remains limited to occlusion alone, neglecting other critical complexity factors.

{We propose MonoDLGD, a novel framework that leverages rich geometric information from ground-truth labels and explicitly models instance-level detection difficulty. Fig.~\ref{fig:MonoDLGD_overview} illustrates the overall architecture of MonoDLGD, consisting of a backbone, 3D detection encoder, and 3D detection decoder. 
Following prior DETR-based architectures~\cite{monodgp,monodetr}, the encoder layer is composed of self-attention layers followed by feedforward layers, while the decoder conducts self-attention across queries and cross-attention between encoder-generated features and queries.}

{MonoDLGD adopts a two-stage architecture utilizing the 3D detection decoder. In Stage 1, label queries are passed through the decoder and two prediction heads to estimate detection uncertainty for projected bounding boxes and depth attributes. Based on the estimated uncertainty, Difficulty-Aware Perturbations (DAP) are applied to generate perturbed label queries. 

This strategy facilitates robust learning by adapting perturbation strength to instance-level difficulty. 
In Stage 2, the perturbed label queries from Stage 1 and the 3D-DAB queries, which explicitly embed spatial priors, are jointly fed to the decoder. The decoder simultaneously performs perturbation reconstruction and 3D object prediction, effectively leveraging instance difficulty and geometric priors to enhance monocular 3D detection.}

\subsection{3D-Dynamic Anchor Box (3D-DAB)} 
\label{subsec:3D-DAB}

{MonoDLGD initializes queries in the detection decoder as 3D-DAB, which explicitly encodes spatial priors rather than using arbitrary learnable embeddings. 
Inspired by DAB-DETR~\cite{liu2022dab}, our 3D-DAB extends dynamic anchor boxes to monocular 3D detection by incorporating projected geometry and class semantics. Each query in the 3D-DAB set $Q_{DAB}=\{q_i|i=1,...,N\}$, where $N$ denotes the number of 3D-DAB queries in a mini-batch, is defined as
\begin{equation}
    q_i = [b^{proj}_{i}, d_i,c_i] \in \mathbb{R}^{7+C},
\end{equation}
where $b^{proj}=[x^{proj}, y^{proj},o^l,o^t,o^r,o^b]\in \mathbb{R}^6$ denotes the bounding box projected onto the normalized 2D image plane, $d\in\mathbb{R}$ is the depth, and $c\in \mathbb{R}^C$ is the class embedding for $C$ object categories. The projected bounding box $b^{proj}$ consists of the center coordinates $(x^{proj}, y^{proj})$ and the distances $(o^l,o^t,o^r,o^b)$ from this center to each of the four sides. }
Here, the superscripts $l,t,r,b$ correspond to the direction from the projected center to each side (left, top, right, bottom) of the bounding box.

{By directly encoding the geometric correspondence between the 2D image plane and 3D object space through projected bounding boxes, 3D-DAB constrains the search space to geometrically meaningful regions, rather than relying on arbitrary learnable embeddings. This significantly reduces detection ambiguity and facilitates more accurate and robust monocular 3D object detection~\cite{liu2022dab}. By embedding these explicit spatial priors into the query representation, 3D-DAB enables the model to localize objects in 3D space more effectively.}

\subsection{Difficulty-Aware Perturbation (DAP)}\label{subsec:DAP}

\begin{figure}[t]
    \centering
    \includegraphics[width=0.72\linewidth]{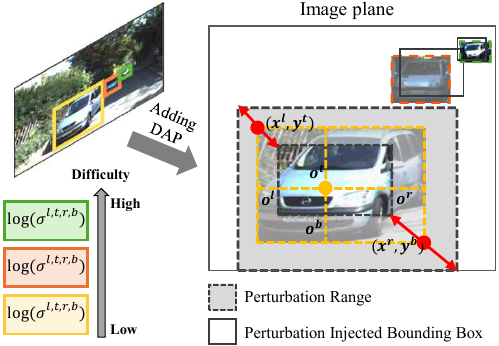}
    \caption{Difficulty-Aware Perturbation (DAP) for projected bounding boxes. Objects with lower uncertainty (lower difficulty scores) receives larger perturbations (\eg, yellow box).}
    \label{fig:perturbation}
\end{figure}

{To address the limited geometric cues and diverse instance difficulty in monocular 3D detection, MonoDLGD introduces the DAP strategy that computes instance-wise difficulty scores based on detector-estimated uncertainty and adaptively scales the perturbation strength for each label query, as shown in Fig.~\ref{fig:MonoDLGD_overview}. Harder objects with higher uncertainty receive smaller perturbations to preserve essential geometric information, while easier objects are perturbed more strongly to effectively regularize training. This results in difficulty-adaptive perturbed label queries, which explicitly guide the model to learn geometry-aware representations through a reconstruction process. Since perturbation and reconstruction are applied only during training, DAP introduces negligible additional inference-time cost.

The perturbed label query set is denoted by $Q_{LP}=\{ \tilde{q}_i| i=1,...,K\}$, where $K$ denotes the number of objects in a mini-batch. The proposed DAP consists of two stages: (i) Difficulty score estimation and (ii) Difficulty-aware label perturbation.}

\subsubsection{Difficulty Score Estimation}

Difficulty scores are computed based on the estimation uncertainty of depth and projected bounding box for each instance. 
We estimate uncertainty using ground-truth labels rather than label queries corrupted with uniform noise, which often require careful hyperparameter tuning and may lead to unstable training dynamics.
Ground-truth labels inherently encode object-level geometry and supervision fidelity, offering a more stable and reliable signal for uncertainty estimation. A detailed comparison with uniformly noised label queries is presented in the supplementary material.

Specifically, as shown in Fig.~\ref{fig:MonoDLGD_overview}, label queries, which consist of the projected bounding box coordinates $b^{proj}$, depth, and one-hot class vector are first processed by the 3D detection decoder in Stage 1. The resulting features are then fed into two separate detection heads, the projected bounding box head and depth head, to estimate the log-variance uncertainties $\text{log}(\sigma^{v})$, where $v \in \{ d, l, t, r, b\}$. 
The subscript $i$ is omitted here for notational simplicity.
{Note that the uncertainties of the projected bounding box are estimated for $(x^l, y^t, x^r, y^b)$, where $(x^l, y^t)$ and $(x^r,y^b)$ are the top-left and bottem-right coordinates, instead of $b^{proj}=(x^{proj}, y^{proj},o^l,o^t,o^r,o^b)$ used in 3D-DAB. This is because the latter would require additional uncertainty estimation for the projection center $x^{proj}, y^{proj}$, complicating the perturbation design.}

To convert uncertainty into a certainty score, we compute the inverse of the log-variance:
\begin{equation}
c^v = \exp(-\log(\sigma^v)),\quad v\in \{ d, l, t, r, b\}.
\end{equation}
The resulting certainty values are then min-max normalized to obtain a relative difficulty score $\hat{c} \in [0,1]$:
\begin{equation}
\hat{c}^v = \frac{ c^v-c^v_\text{min} }{ c^v_\text{max}-c^v_\text{min} },
\end{equation}
where $c^v_\text{min}$ and $c^v_\text{max}$ represent the minimum and maximum of the certainty $c$ over entire training dataset.
A higher $\hat{c}$ indicates greater prediction certainty. To ensure that normalization captures the global distribution of prediction difficulties throughout training, the minimum and maximum certainty values are updated at each batch using an exponential moving average (EMA):
\begin{equation}
    c^v_{min,t} \leftarrow \beta c^v_{min, t-1} + (1-\beta) c^v_{min, t},
\end{equation}
where $c^v_{min,t}$ is the minimum certainty at iteration $t$, and $\beta$ is the momentum coefficient. $c^v_{max,t}$ is computed in the same way. $c^v_{min,0}$, $c^v_{max,0}$ are initialized from the first mini-batch.

\subsubsection{Difficulty-Aware Label Perturbation}
The computed instance-wise difficulty score $\hat{c}^v_i$ for $v\in \{ d, l, t, r, b\}$ is subsequently used as a perturbation scale factor. This perturbation is independently applied to the depth $d$ and the projected bounding box coordinates $b^{2D}=[x^l, y^t, x^r, y^b]$.

\noindent\textbf{(a) Projected Bounding box Perturbation}: 

Fig.~\ref{fig:perturbation} shows the overall procedure of the bounding box perturbation.
We inject perturbations into the bounding box coordinates $b^{2D}=[x^l, y^t, x^r, y^b]$ using the computed difficulty scores. Specifically, to calculate each coordinates perturbation scale, we independently sample a random sign $s\sim U\{-1,1\}$ and multiply it by the difficulty score $\hat{c} \in[0,1]$, the boundary distance $[o^l,o^t,o^r,o^b]$ and a bounding box perturbation scaling factor $\gamma^{b}\in (0,1)$.
The resulting perturbed coordinates $\tilde{b}^{2D}$ are computed as follows:
\begin{equation}
\begin{aligned}
    \tilde{x}^v &= \text{CLIP}_{(0,1)}(x^v+ o^v \cdot \hat{c}^{v} \cdot s^{v} \cdot \gamma^{b}), \quad v \in \{l,r\}, \\
    \tilde{y}^v &= \text{CLIP}_{(0,1)}(y^v+ o^v \cdot \hat{c}^{v} \cdot s^{v} \cdot \gamma^{b}), \quad v \in \{t,b\}.\\
\end{aligned}
\end{equation}

$\text{CLIP}_{a,b}(x)=min(max(x, a),b)$ ensures that the perturbed bounding box remains within the normalized image plane coordinates range $[0,1]$. 

The perturbation is constrained within the range $-1< \hat{c}^v \cdot s^v \cdot \gamma^{b}<1$, so the perturbed coordinates satisfy $\tilde{x}^l < \tilde{x}^r$ and $\tilde{y}^t < \tilde{y}^b$.
Through this approach, we ensure that the perturbation scale of each $b^{2D}$ remains within the boundary distances and satisfies the geometric constraints $0 \le \tilde{x}^l < \tilde{x}^r \le 1$ and $0 \le \tilde{y}^t < \tilde{y}^b\le 1$ of a valid bounding box. Alg.~\ref{alg:DTA} shows the process of perturbing projected bounding boxes.

\begin{algorithm}[th!]
    \caption{DAP for Projected Bounding Box}
    \label{alg:DTA}
\begin{algorithmic}
    \STATE {\textbf{Input:} $b^{proj}=(x^{proj}, y^{proj},o^l,o^t,o^r,o^b)$, $(\hat{c}^l, \hat{c}^t, \hat{c}^r, \hat{c}^b)$}
    \STATE {\textbf{Reparameterize}} \\
     $b^{proj} \rightarrow {b^{2D}}=(x^l,y^t,x^r,y^b)$ \\
     $x^l=x^{proj}-o^l,\quad y^t=y^{proj}-o^t$ \\
     $x^r=x^{proj}+o^r,\quad y^b=y^{proj}+o^b$

    \STATE {\textbf{Difficulty-Aware Label Perturbation}} \\

    \FOR{each coordinate $v \in \{l,t,r,b\}$}
        \STATE Sample random sign $s^{v} \sim U\{-1,1\}$
        \STATE Compute perturbation scale: $\Delta = o^v \cdot \hat{c}^v \cdot s^v \cdot \gamma^{b}$ 
        % \\ where $o^v \in \{l,t,r,b\}$ denotes the corresponding boundary distance
        \IF{$v \in \{l,r\}$}
            \STATE Update coordinate: $\tilde{x}^v = \text{CLIP}_{(0,1)}(x^v + \Delta)$
        \ELSIF{$v \in \{t,b\}$}
            \STATE Update coordinate: $\tilde{y}^v = \text{CLIP}_{(0,1)}(y^v + \Delta)$        
        \ENDIF
    \ENDFOR

    \STATE Obtain perturbed box: $\tilde{b}^{2D}=(\tilde{x}^l,\tilde{y}^t,\tilde{x}^r,\tilde{y}^b)$

    \STATE 
    {\textbf{Inverse-Reparameterize} } \\
    $ \tilde{b}^{2D}\rightarrow \tilde{b}^{proj}=(\tilde{x}^{proj}, \tilde{y}^{proj},\tilde{o}^l,\tilde{o}^t,\tilde{o}^r,\tilde{o}^b), $ \\
    where $(\tilde{x}^{proj}, \tilde{y}^{proj})$ are the center coordinates of the \emph{reparameterized} perturbed bounding box $\tilde{b}^{2D}$.

    \STATE \textbf{Output:} Perturbed projected bounding box $\tilde{b}^{proj}$
    
\end{algorithmic}
\end{algorithm}

\noindent\textbf{(b) Depth Perturbation}:
The depth perturbation is performed similarly to the bounding box perturbation. To determine the depth perturbation scale, we multiply the depth $d$ by a randomly sampled sign $s^{d}\sim U\{-1,1\}$, the depth difficulty score $\hat{c}^d\in(0,1)$, and a depth perturbation scaling factor $\gamma^d\in(0,1)$ as follows:
\begin{equation}
    \tilde{d} = d + d\cdot \hat{c}^d \cdot s^{d} \cdot \gamma^d.
\end{equation}

\noindent\textbf{(c) Class Perturbation}:
{In monocular 3D object detection, class information serves as a strong prior to constraining object size and aspect ratio. Therefore, perturbing the class label during training can act as a useful form of regularization. We adopt a label-flipping strategy where class labels are randomly switched to another class with equal probability. Unlike depth or bounding box perturbations, class perturbation is difficulty-agnostic and applied uniformly across all instances. %, regardless of their predicted uncertainty.
The final perturbed label query is defined as:}
\begin{equation}
    \tilde{q}_i=[\tilde{b}_i, \tilde{d}_i, \tilde{c}_i].
\end{equation}

\subsection{Difficulty-Aware Reconstruction}
The perturbed label queries $Q_{LP}=\{\tilde{q_i}|i=1,...,K\}$, generated through DAP, are fed into the 3D detection decoder along with the 3D-DAB queries $Q_{DAB}=\{q_i|i=1,...,N\}$. $K$ and $N$ are the number of objects in a mini-batch and the number of 3D anchor queries, respectively. Both query sets share the same detection heads (projected Bbox, depth, category), as shown in Fig.~\ref{fig:MonoDLGD_overview}. 
Additionally, both the projected bounding box head and the depth head estimate the uncertainty of each corresponding attribute.

The decoder is supervised by (i) reconstructing original labels from $Q_{LP}$ and (ii) detecting objects from $Q_{DAB}$. Since the perturbed label query $\tilde{q_i}$ has a known corresponding ground truth label, Hungarian matching is not required for reconstruction loss ($L_{recon}$).
For reconstructing the projected bounding box and depth, we employ the Laplacian aleatoric uncertainty loss, enabling uncertainty-adaptive training:
\begin{equation}
    L^d_{recon}= \sum_{i=1}^{K} \Bigg( \frac{\sqrt{2}}{\sigma^d_i}|| d_{gt,i}-d_{recon,i}||_1 + log(\sigma^d_i) \Bigg),
\end{equation}

\begin{align}
    L^{bbox}_{recon} = \sum_{i=1}^{K} \Bigg(  & \sum_{v \in \{l, r\}} ( \frac{\sqrt{2}}{\sigma^v_i} \left\| x^v_{gt,i} - x^v_{recon,i} \right\|_1 + \log(\sigma^v_i) ) \nonumber\\
        + & \sum_{v \in \{t, b\}} ( \frac{\sqrt{2}}{\sigma^v_i} \left\| y^v_{gt,i} - y^v_{recon,i} \right\|_1 + \log(\sigma^v_i) )
    \Bigg)
\end{align}

\noindent where $(x_{gt},y_{gt})$ and $(x_{recon},y_{recon})$ represent ground truth and reconstructed coordinates of \emph{reparameterized} projected bounding boxes, respectively. Class reconstruction utilizes standard cross-entropy loss.

The overall reconstruction loss is defined as:
\begin{equation}
    L_{recon} = \lambda_{bbox}L^{bbox}_{recon} + \lambda_{d}L^{d}_{recon} + \lambda_{cls}L^{cls}_{recon},
\end{equation}
where $\lambda$ denotes the weighting factor to balance losses.
Our proposed module can be easily integrated as a plug-in into existing DETR-based 3D object detectors, introducing negligible additional inference-time computational overhead as perturbation and reconstruction occur only during training.

\subsection{Loss Function}

The overall training objective comprises the label reconstruction loss $L_{recon}$ and the detection loss $L_{det}$ adopted from baseline methods.
For predictions from the 3D-DAB queries, we perform Hungarian matching with ground truth labels, and then apply the same loss function as MonoDGP~\cite{monodgp} for orientation, 3D size, 2D projected bounding box, depth, and class:
\begin{equation}
    L = L_{recon}+L_{det}.
\end{equation}

\begin{table*}[th!]
\caption{\textbf{Comparisons on the KITTI test and validation sets (Car category)}. We \textbf{bold} the best results and \underline{underline} the second-best results.}
\centering
\begingroup
\renewcommand{\arraystretch}{0.9}
\scriptsize  
\setlength{\tabcolsep}{1mm}
\begin{tabular}{@{}l|c|c|ccc|clcc|clcc|clcc@{}}
\hline
\multicolumn{1}{c|}{} & & & \multicolumn{3}{c|}{Test, $AP_{BEV|R40}$} & \multicolumn{4}{c|}{Test, $AP_{3D|R40}$} & \multicolumn{4}{c|}{Val, $AP_{BEV|R40}$} & \multicolumn{4}{c}{Val, $AP_{3D|R40}$} \\  
\multicolumn{1}{l|}{\multirow{-2}{*}{Methods}} & \multirow{-2}{*}{Extra data} & \multirow{-2}{*}{Reference} & Easy & Mod. & Hard & \multicolumn{2}{c@{}}{Easy} & Mod. & Hard & \multicolumn{2}{c@{}}{Easy} & Mod. & Hard & \multicolumn{2}{c@{}}{Easy} & Mod. & Hard \\ \hline
%CaDDN~\cite{caddn} &  & CVPR 2021 & 27.94 & 18.91 & 17.19 & \multicolumn{2}{c}{19.17} & 13.41 & 11.46 & \multicolumn{2}{c}{-} & - & - & \multicolumn{2}{c}{23.57} & 16.31 & 13.84 \\
MonoDTR~\cite{monodtr} & & CVPR 2022 & 28.59 & 20.38 & 17.14 & \multicolumn{2}{c}{21.99} & 15.39 & 12.73 & \multicolumn{2}{c}{33.33} & 25.35 & 21.68 & \multicolumn{2}{c}{24.52} & 18.57 & 15.51 \\
DID-M3D~\cite{DID-M3D} & & ECCV 2022 & 32.95 & 22.76 & 19.83 & \multicolumn{2}{c}{24.40} & 16.29 & 13.75 & \multicolumn{2}{c}{31.10} & 22.76 & 19.50 & \multicolumn{2}{c}{22.98} & 16.12 & 14.03 \\
OccupancyM3D~\cite{occupancym3d} & \multirow{-4}{*}{LiDAR} & CVPR 2024 & 35.38 & 24.18 & 21.37 & \multicolumn{2}{c}{25.55} & 17.02 & 14.79 & \multicolumn{2}{c}{35.72} & 26.60 & 23.68 & \multicolumn{2}{c}{26.87} & 19.96 & 17.15 \\ \hline
MonoPGC~\cite{monopgc} & & ICRA 2023 & 32.50 & 23.14 & 20.30 & \multicolumn{2}{c}{24.68} & 17.17 & 14.14 & \multicolumn{2}{c}{34.06} & 24.26 & 20.78 & \multicolumn{2}{c}{25.67} & 18.63 & 15.65 \\
OPA-3D~\cite{opa-3d} & \multirow{-2}{*}{Depth} & RAL 2023 & 33.54 & 22.53 & 19.22 & \multicolumn{2}{c}{24.60} & 17.05 & 14.25 & \multicolumn{2}{c}{33.80} & 25.51 & 22.13 & \multicolumn{2}{c}{24.97} & 19.40 & 16.59 \\ \hline
%GUPNet~\cite{gupnet} &  & ICCV 2021 & - & - & - & \multicolumn{2}{c}{20.11} & 14.20 & 11.77 & \multicolumn{2}{c}{31.07} & 22.94 & 19.75 & \multicolumn{2}{c}{22.76} & 16.46 & 13.72 \\
%MonoCon~\cite{monocon} &  & AAAI 2022 & 31.12 & 22.10 & 19.00 & \multicolumn{2}{c}{22.50} & 16.46 & 13.95 & \multicolumn{2}{c}{-} & - & - & \multicolumn{2}{c}{26.33} & 19.01 & 15.98 \\
DEVIANT~\cite{deviant} & & ECCV 2022 & 29.65 & 20.44 & 17.43 & \multicolumn{2}{c}{21.88} & 14.46 & 11.89 & \multicolumn{2}{c}{32.60} & 23.04 & 19.99 & \multicolumn{2}{c}{24.63} & 16.54 & 14.52 \\
MonoDDE~\cite{monodde} & & CVPR 2022 & 33.58 & 23.46 & 20.37 & \multicolumn{2}{c}{24.93} & 17.14 & 15.10 & \multicolumn{2}{c}{35.51} & 26.48 & 23.07 & \multicolumn{2}{c}{26.66} & 19.75 & 16.72 \\
MonoUNI~\cite{monouni} & & NeurlPS 2023 & - & - & - & \multicolumn{2}{c}{24.75} & 16.73 & 13.49 & \multicolumn{2}{c}{-} & - & - & \multicolumn{2}{c}{24.51} & 17.18 & 14.01 \\
MonoDETR~\cite{monodetr} & & ICCV 2023 & 33.60 & 22.11 & 18.60 & \multicolumn{2}{c}{25.00} & 16.47 & 13.58 & \multicolumn{2}{c}{37.86} & 26.95 & 22.80 & \multicolumn{2}{c}{28.84} & 20.61 & 16.38 \\
MonoCD~\cite{monocd} & & CVPR 2024 & 33.41 & 22.81 & 19.57 & \multicolumn{2}{c}{25.53} & 16.59 & 14.53 & \multicolumn{2}{c}{34.60} & 24.96 & 21.51 & \multicolumn{2}{c}{26.45} & 19.37 & 16.38 \\
FD3D~\cite{fd3d} & \multirow{-8}{*}{None} & AAAI 2024 & 34.20 & 23.72 & 20.76 & \multicolumn{2}{c}{25.38} & 17.12 & 14.50 & \multicolumn{2}{c}{36.98} & 26.77 & 23.16 & \multicolumn{2}{c}{28.22} & 20.23 & 17.04 \\ 

MonoMAE~\cite{monomae}& & NeurlPS 2024 & 34.15 & 24.93 & 21.76 & \multicolumn{2}{c}{25.60} & \underline{18.84} & \underline{16.78} & \multicolumn{2}{c}{\underline{40.26}} & 27.08 & 23.14 & \multicolumn{2}{c}{30.29} & 20.90 & 17.61 \\ 

MonoDGP~\cite{monodgp} & & CVPR 2025 & \underline{35.24} & \underline{25.23} & \underline{22.02} & \multicolumn{2}{c}{\underline{26.35}} & 18.72 & 15.97 & \multicolumn{2}{c}{39.40} & \underline{28.20} & \underline{24.42} & \multicolumn{2}{c}{\underline{30.76}} & \underline{22.34} & \underline{19.02} \\ \hline

Ours & None & & \textbf{36.63} & \textbf{25.3} & \textbf{23.13} & \multicolumn{2}{c}{\textbf{29.11}} & \textbf{19.87} & \textbf{17.74} & \multicolumn{2}{c}{\textbf{41.68}} & \textbf{30.53} & \textbf{27.76} & \multicolumn{2}{c}{\textbf{34.89}} & \textbf{25.19} & \textbf{21.78} \\
Improvement over Second-Best Method & - & & +1.39 & +0.07 & +1.11 & \multicolumn{2}{c}{+2.76} & +1.03 & +0.96 & \multicolumn{2}{c}{+1.42} & +2.33 & +3.34 & \multicolumn{2}{c}{+4.13} & +2.85 & +2.76 \\ 

Improvement over MonoDGP Baseline & - & & +1.39 & +0.07 & +1.11 & \multicolumn{2}{c}{+2.76} & +1.15 & +1.77 & \multicolumn{2}{c}{+2.28} & +2.33 & +3.34 & \multicolumn{2}{c}{+4.13} & +2.85 & +2.76 \\ 
\hline
\end{tabular}
\endgroup
\label{Table:main}
\end{table*}

%%%%%%%%%%%%%%%%%%%%%%%%%%%%%%%%%%%%%%%%%%%%%%%
%%              Experiments                  %%  
%%%%%%%%%%%%%%%%%%%%%%%%%%%%%%%%%%%%%%%%%%%%%%%

\section{Experiments}

\subsection{Experimental Setup}
\paragraph{Dataset}
We evaluated our method on the KITTI 3D object detection benchmark~\cite{kitti}, a widely used dataset in monocular 3D detection. The dataset contains 7,481 training images and 7,518 testing images, and provides annotations for three object categories (Car, Pedestrian, and Cyclist). Each object is further classified into three difficulty levels (Easy, Moderate, Hard). Following the common protocol established in~\cite{chen20153d}, we split the 7,481 training images into 3,712 for training and 3,769 for validation.

\paragraph{Evaluation Metrics}
We reported results across the three difficulty levels (Easy, Moderate, Hard) using Average Precision (AP) for 3D bounding boxes \( \mathrm{AP}_{3D} \) and bird eye view projection \( \mathrm{AP}_{BEV} \). These metrics are computed over 40 recall positions in accordance with the official KITTI protocol~\cite{simonelli2019disentangling}. All methods were ranked based on the Moderate \( \mathrm{AP}_{3D} \) score for the Car category.

\paragraph{Implementation Details}

Our method is implemented on top of MonoDGP~\cite{monodgp}, which employs ResNet-50~\cite{resnet} as the backbone network.
The loss for detection $L_{det}$ follow the MonoDGP setup.
We conducted both main and ablation experiments based on the MonoDGP-based implementation. The model was trained for 250 epochs using the Mixup3D~\cite{monolss} strategy following~\cite{monodgp}. The batch size and initial learning rate were set to 8 and \( 2 \times 10^{-4} \), respectively. The AdamW optimizer~\cite{adamW} was used with a weight decay of \( 10^{-4} \) and the learning rate decayed by a factor of 0.5 at epochs 85, 125, 165, and 225. 
Training was conducted on an NVIDIA RTX A6000.
During inference, we discarded queries with category confidence scores fall below 0.2.
In addition, we applied our method to a MonoDETR-based implementation. Training follows the same setup as MonoDGP.

%%%%%%%%%%%%%%%%%%%%%%%%%%%%%%%%%%%%%%%%%%%%%%%%%%%%%%%%%%%%%%%%%%%%%%%%%%%%%%%%%%
\begin{table}
\caption{
\textbf{More results with computational cost.} The proposed method was implemented on top of MonoDETR~\cite{monodetr} and MonoDGP~\cite{monodgp}. * indicates results reproduced from the authors' official code. Performance was evaluated on the KITTI validation set for Car category.
}
\centering
\begingroup
\scriptsize
\setlength{\tabcolsep}{1mm}
\renewcommand{\arraystretch}{0.5}
\begin{tabular}{l|clcc|clcc|cc}
\toprule
\multicolumn{1}{l|}{\multirow{2}{*}{Method}} 
& \multicolumn{4}{c|}{Val, $AP_{BEV|R40}$} 
& \multicolumn{4}{c|}{Val, $AP_{3D|R40}$} 
& \multirow{2}{*}{\shortstack{GFLOPs$\downarrow$}} 
& \multirow{2}{*}{\shortstack{Time\\(ms)$\downarrow$}}  % Runtime->Time 변경
\\ 
& \multicolumn{2}{c}{Easy} & Mod. & Hard 
& \multicolumn{2}{c}{Easy} & Mod. & Hard 
\\ \midrule

MonoDETR*
& \multicolumn{2}{c}{36.38} & 26.19 & 22.29 
& \multicolumn{2}{c}{27.34} & 19.33 & 16.04 
& 59.7 & 35.2 \\

+Ours 
& \multicolumn{2}{c}{38.59} & 27.65 &  23.62
& \multicolumn{2}{c}{29.79} & 21.63 & 18.17 
& 59.8 & 35.5 \\

\midrule
MonoDGP
& \multicolumn{2}{c}{39.40} & 28.20 &  24.42
& \multicolumn{2}{c}{30.76} & 22.34 & 19.02 
& 69.0 & 42.4 \\
+Ours
& \multicolumn{2}{c}{41.68} & 30.53 &  27.76
& \multicolumn{2}{c}{34.89} & 25.19 & 21.78
& 69.3 & 42.7 \\

\bottomrule
\end{tabular}

\endgroup
\label{Table:monodetr}
\end{table}

\subsection{Main Results}

We evaluated our proposed method, implemented on top of the MonoDGP~\cite{monodgp} architecture. Table~\ref{Table:main} reports results on the KITTI 3D test set, evaluated using the official online server~\cite{kitti} for fair comparison.
To facilitate better learning of 3D geometric information from objects with varying levels of difficulty, our method introduces a difficulty-aware label-guided denoising strategy. As shown, MonoDLGD achieves state-of-the-art performance across all difficulty levels, without relying on any additional training data. 
Compared to the MonoDGP baseline, MonoDLGD significantly improves \( \mathrm{AP}_{3D}^{R40} \) by +2.76 (Easy), +1.15 (Moderate), and +1.77 (Hard) on the test set. These consistent gains demonstrate that MonoDLGD effectively enhances 3D geometric reasoning in monocular settings, especially under challenging conditions such as occlusion, truncation, and depth ambiguity.

To further evaluate the versatility of our approach, we integrated MonoDLGD into the MonoDETR~\cite{monodetr} architecture without modifying its core design. Specifically, we applied our label denoising strategy in conjunction with 3D-DAB queries. As shown in Table~\ref{Table:monodetr}, this integration yields consistent improvements. These results suggest that our method can serve as a complementary component to existing DETR-based monocular 3D detection pipelines, potentially benefiting a broader range of architectures with similar formulations. 

\subsection{Ablation Study}

\begin{table}[t!]
\caption{
\textbf{Ablation on the KITTI val set (Car)}. \textbf{(a)} serves as our baseline with the same architecture as MonoDGP~\cite{monodgp}. \textbf{UN} indicates uniform noise ignores instance-level detection difficulty, equivalent to DN-DETR~\cite{dndetr}. \textbf{$L_{recon}^{bbox}$} means the type of projected bounding box reconstruction loss. \textbf{LU} is Laplacian Uncertainty Loss adopted in our method.}
\centering
\begingroup
\scriptsize
\setlength{\tabcolsep}{1.8pt}
\renewcommand{\arraystretch}{0.5}
\begin{tabular}{c |c|c|c| clcc|clcc}
\toprule
\multirow{2}{*}{Idx}
% &\multicolumn{3}{c|}{Setting} 
% \multirow{2}{*}{\shortstack{Flops\\(G)$\downarrow$}} 
&\multirow{2}{*}{\shortstack{3D-DAB}}
&\multirow{2}{*}{Perturb.}
&\multirow{2}{*}{$L_{recon}^{bbox}$}
& \multicolumn{4}{c|}{Val, $AP_{BEV|R40}$} 
& \multicolumn{4}{c}{Val, $AP_{3D|R40}$} 
\\ 
& & &
& \multicolumn{2}{c}{Easy} & Mod. & Hard 
& \multicolumn{2}{c}{Easy} & Mod. & Hard 
\\ \midrule

(a) & $\times$ & $\times$ & $\times$ 
& \multicolumn{2}{c}{39.40} & 28.20 & 24.42 
& \multicolumn{2}{c}{30.76} & 22.34 & 19.02 \\ \midrule

(b) & O & $\times$ & $\times$
& \multicolumn{2}{c}{36.85} & 26.72 &  23.21
& \multicolumn{2}{c}{27.82} & 20.64 & 17.78 \\

(c) & O & UN & L1
& \multicolumn{2}{c}{40.32} & 30.13 & 26.53
& \multicolumn{2}{c}{31.99} & 23.82 & 20.65 \\

(d) & O & UN & LU
& \multicolumn{2}{c}{41.16} & 30.31 & 26.54 
& \multicolumn{2}{c}{33.82} & 24.7 & 21.19 \\

(e): Ours & O & DAP & LU
& \multicolumn{2}{c}{41.68} & 30.53 & 27.76
& \multicolumn{2}{c}{34.89} & 25.19 & 21.78 \\

\bottomrule
\end{tabular}

\endgroup
\label{Table:ablation}
\end{table}

%%%%%%%%%%%%%%%%%%%%%%%%%%%%%%%%%%%%%%%%%%%%%%%%%%%%%%%%%%%%%%%%%%%%%

\paragraph{Efficiency Comparison} 
Table~\ref{Table:monodetr} compares the inference time on the KITTI validation set. All methods were evaluated under the same computational environment using a single NVIDIA Titan RTX GPU with a batch size of 1 to ensure fair comparison. The average inference time per image for MonoDGP~\cite{monodgp} and MonoDETR~\cite{monodetr} is 42.4 ms and 35.2 ms, respectively. When integrated with the proposed method, the inference time remains nearly unchanged. 
The training time increases slightly due to the added perturbation and reconstruction operations in Stage 1 (see Fig.~\ref{fig:MonoDLGD_overview}), which are applied only during training. A detailed analysis is provided in the supplementary.

\paragraph{Effectiveness of Label-Guided Denoising with 3D-DAB and DAP}
Table~\ref{Table:ablation} presents an ablation study evaluating the core components of MonoDLGD framework. 
Starting from the MonoDGP baseline (a), simply replacing anchor queries 3D-DAB (b) slightly degrades performance due to the lack of accompanying supervision despite encoding spatial priors. However, combining 3D-DAB with uniform label perturbation (c) yields immediate performance gains, demonstrating that label-guided denoising enhances 3D geometric learning by providing explicit supervision to the anchor queries. 
Further gains are achieved by introducing Difficulty-Aware Perturbation (DAP) based on predictive uncertainty, \ie (d) \vs (e).
Unlike the uniform noise scheme of DN-DETR~\cite{dndetr}, DAP adaptively regularizes easy instances while preserving the geometric structure of hard examples. 
Overall, while 3D-DAB provides strong geometric supervision for monocular 3D detection, DAP enables more robust learning by adapting to instance-level difficulty.

\begin{table}[t!]
\caption{\textbf{Ablation study on denoising target configurations} on the KITTI validation set (Car category). 
}
\centering
\begingroup
\scriptsize
\setlength{\tabcolsep}{1mm}
\renewcommand{\arraystretch}{0.4}
\begin{tabular}{l|clcc|clcc}
\toprule
\multicolumn{1}{l|}{\multirow{2}{*}{Denoising Setup}} 
& \multicolumn{4}{c|}{Val, $AP_{BEV|R40}$} 
& \multicolumn{4}{c}{Val, $AP_{3D|R40}$} 
\\ 
& \multicolumn{2}{c}{Easy} & Mod. & Hard 
& \multicolumn{2}{c}{Easy} & Mod. & Hard 
\\ \midrule

 Baseline (MonoDGP)
& \multicolumn{2}{c}{39.40} & 28.20 & 24.42 
& \multicolumn{2}{c}{30.76} & 22.34 & 19.02 \\

\midrule
% Bbox (proj.) + Cls 
Bbox (proj.) + Class
& \multicolumn{2}{c}{40.58} & 29.75 &  26.18
& \multicolumn{2}{c}{30.93} & 23.36 & 20.30 \\
% Bbox (proj.) + Cls + Depth
Bbox (proj.) + Class + Depth
& \multicolumn{2}{c}{41.68} & 30.53 & 27.76
& \multicolumn{2}{c}{34.89} & 25.19 & 21.78 \\
\bottomrule
\end{tabular}
\endgroup
\label{Table:ablation2}
\end{table}

\paragraph{Effectiveness of Uncertainty in Denoising}
We extend the use of aleatoric uncertainty to the bounding box denoising process.
As shown in Table~\ref{Table:ablation}, this extension yields a notable performance improvement, \ie (c) \vs (d). 
Incorporating uncertainty-aware estimation into the denoising process helps the model downweight unreliable supervision from hard or ambiguous objects, allowing it to focus on more confident and informative signals.
These results suggest that aleatoric uncertainty provides a meaningful signal for modeling instance-level detection difficulty in label denoising.

\paragraph{Effectiveness of Depth Information}
To evaluate the contribution of depth information in our method, we conducted an ablation study by excluding depth attributes from both the label queries and the denoising process. As shown in Table~\ref{Table:ablation2}, applying our denoising strategy solely to the projected bounding boxes and class labels improves the moderate \( \mathrm{AP}_{3D}^{R40} \) from 22.34\% to 23.36\%, validating the effectiveness of denoising core detection components. More importantly, further incorporating depth into the denoising process yields a significant performance boost to 25.19\%, highlighting the importance of depth supervision in enhancing geometric representation.

\section{Conclusion}

We have presented MonoDLGD, a novel framework for monocular 3D object detection that introduces difficulty-aware label-guided denoising via label perturbation and reconstruction. By adaptively modulating perturbation strength based on reconstruction uncertainty, our method explicitly incorporates geometric supervision and effectively mitigates the ill-posed nature of monocular 3D object detection. Furthermore, our uncertainty-aware estimation strategy leads to consistent performance gains, highlighting the importance of modeling instance-level uncertainty. Extensive experiments on the KITTI benchmark demonstrate that MonoDLGD consistently improves 3D detection performance across all difficulty levels.

\section{Acknowledgments}
This work was supported by the NRF of Korea grant (RS-2025-24803204), the Bio \& Medical Technology Development Program of the NRF of Korea (RS-2022-NR068424), and the BK21 FOUR funded by the Ministry of Education (MOE, Korea) and the NRF of Korea (2120251015523).

% \bigskip
% \noindent Thank you for reading these instructions carefully. We look forward to receiving your electronic files!

\bibliography{aaai2026}

\clearpage

\section{Appendix}

\subsection{Analyzing the Relationship Between Uncertainty and Difficulty} 
The proposed Difficulty-Aware Perturbation (DAP) module computes a difficulty score for each instance based on the predicted aleatoric uncertainty of its depth and \emph{reparameterized} projected bounding box in Eq. (8) and (9) of the main paper. This score is then used to modulate the perturbation magnitude in Eq. (5) and (6) of the main paper. Instances with higher uncertainty (\ie, greater difficulty) are assigned smaller perturbations to preserve critical geometric cues, and \emph{vice versa}.

\begin{figure*}  
    \centering
  \begin{subfigure}{0.3\linewidth}
    \centering
    \includegraphics[width=\linewidth]{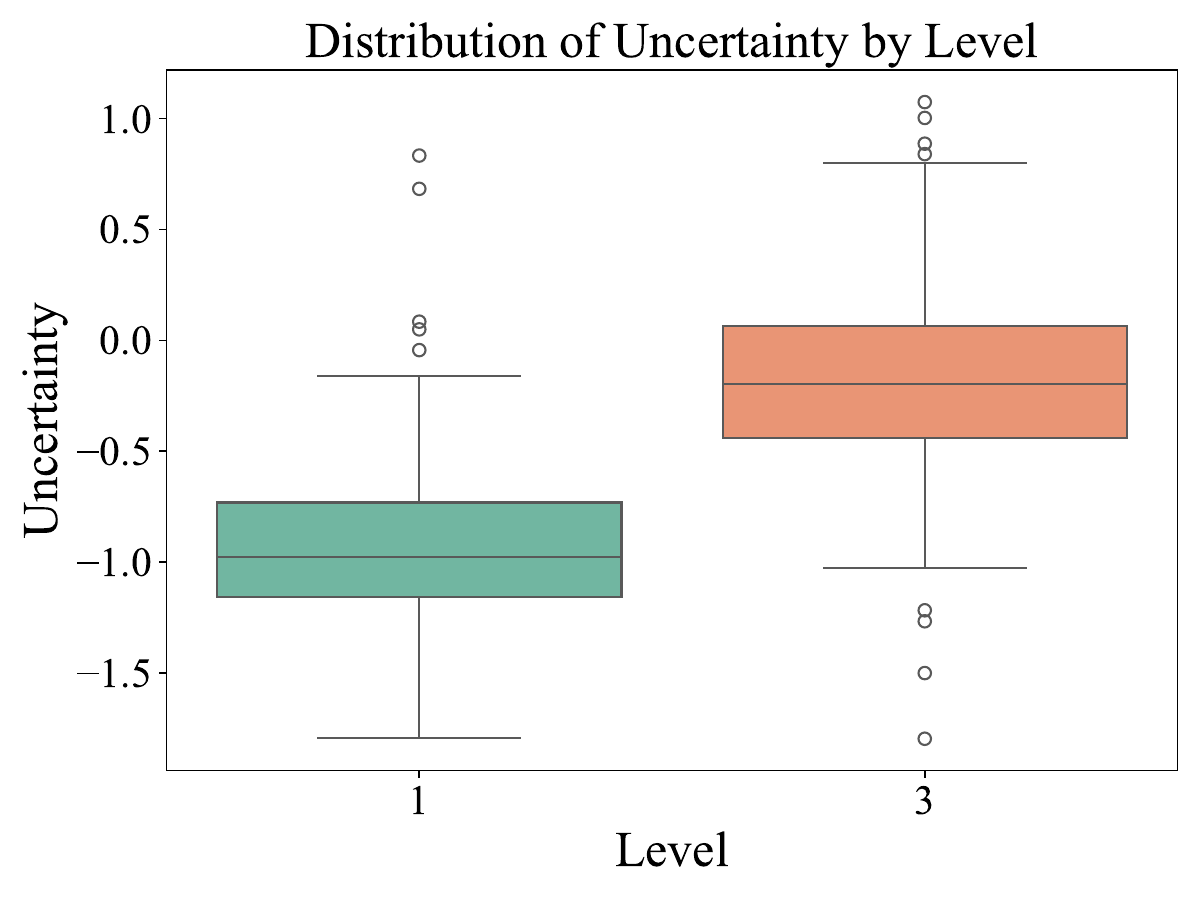}    
    \caption{Depth}
  \end{subfigure} 
  \begin{subfigure}{0.3\linewidth}
    \centering
    \includegraphics[width=\linewidth]{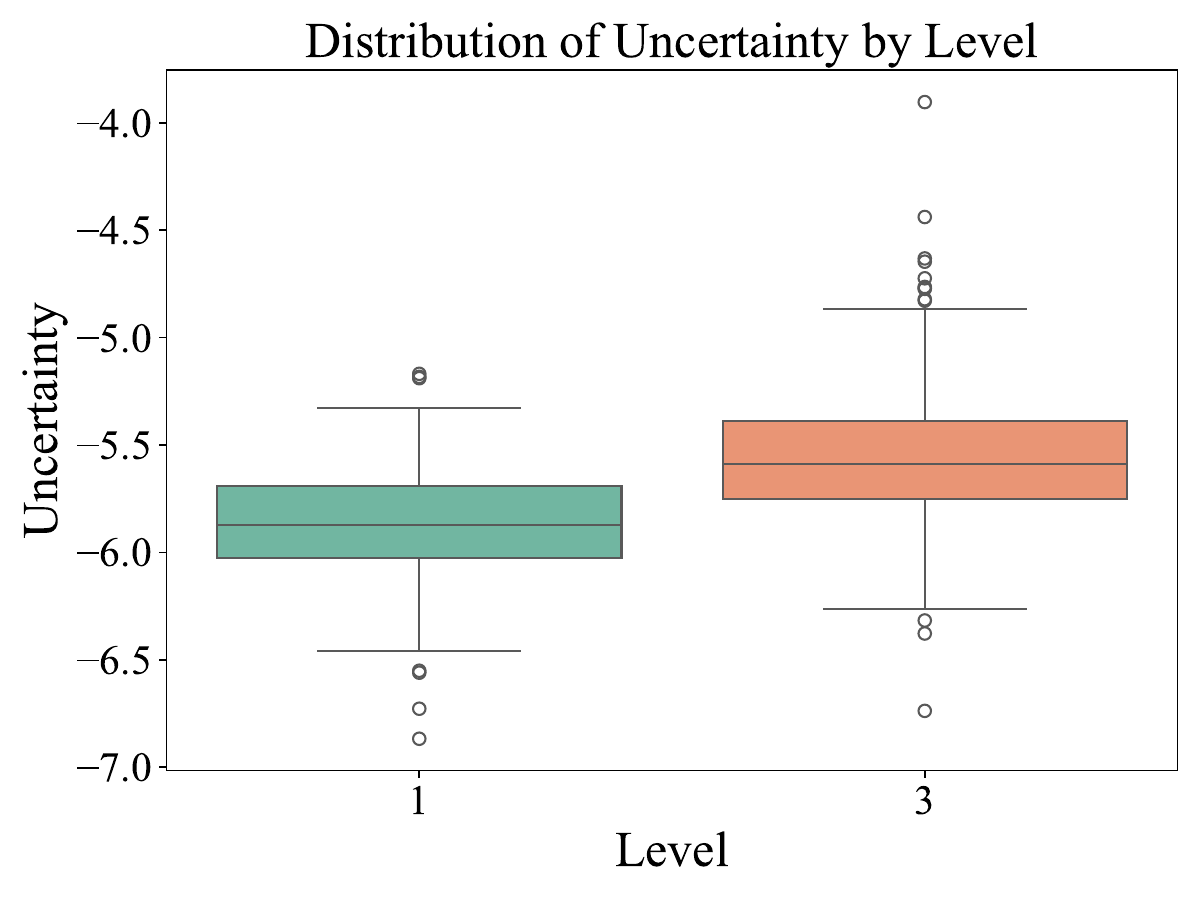} 
    \caption{Projected Bounding Box ($x^l$)}
  \end{subfigure}
  \begin{subfigure}{0.3\linewidth}
    \centering
    \includegraphics[width=\linewidth]{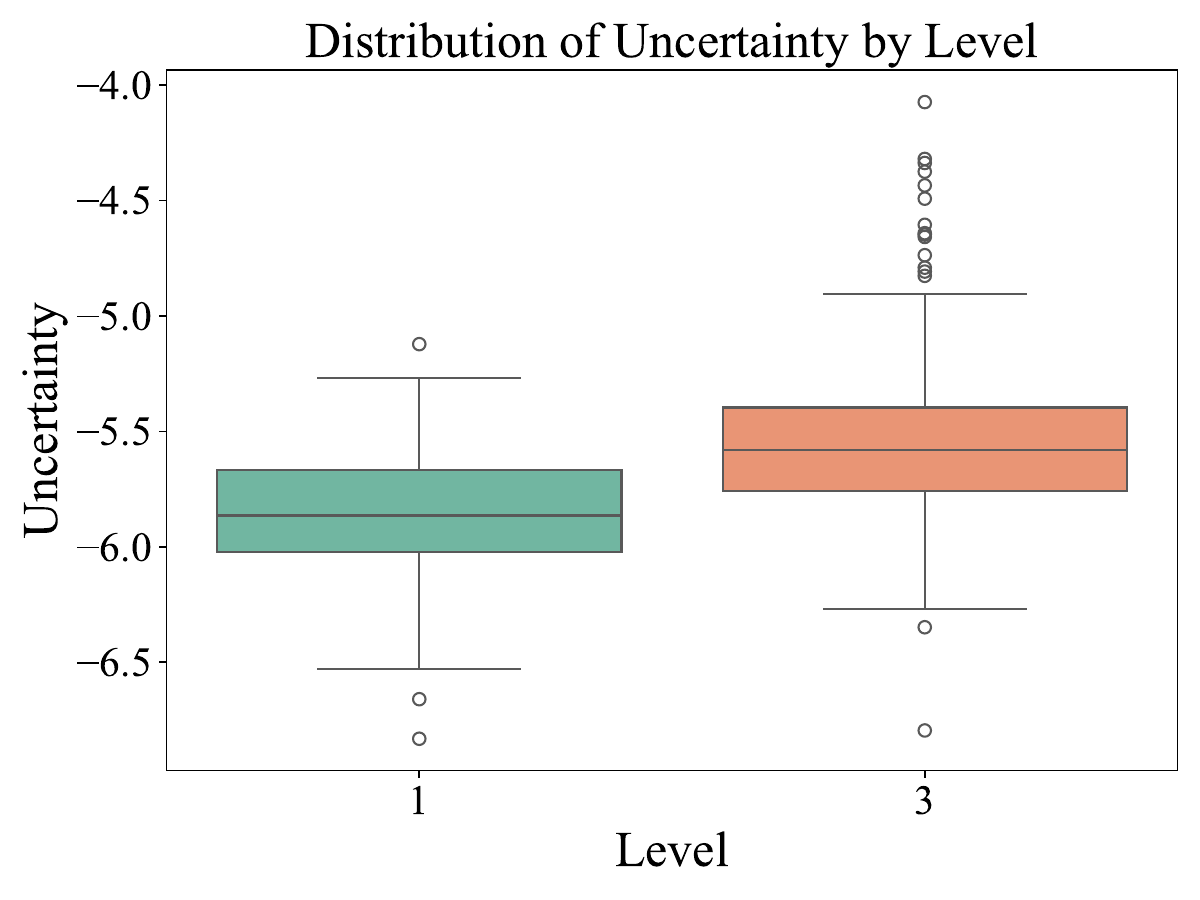} 
    \caption{Projected Bounding Box ($x^r$)}
  \end{subfigure}
  \begin{subfigure}{0.3\linewidth}
    \centering
    \includegraphics[width=\linewidth]{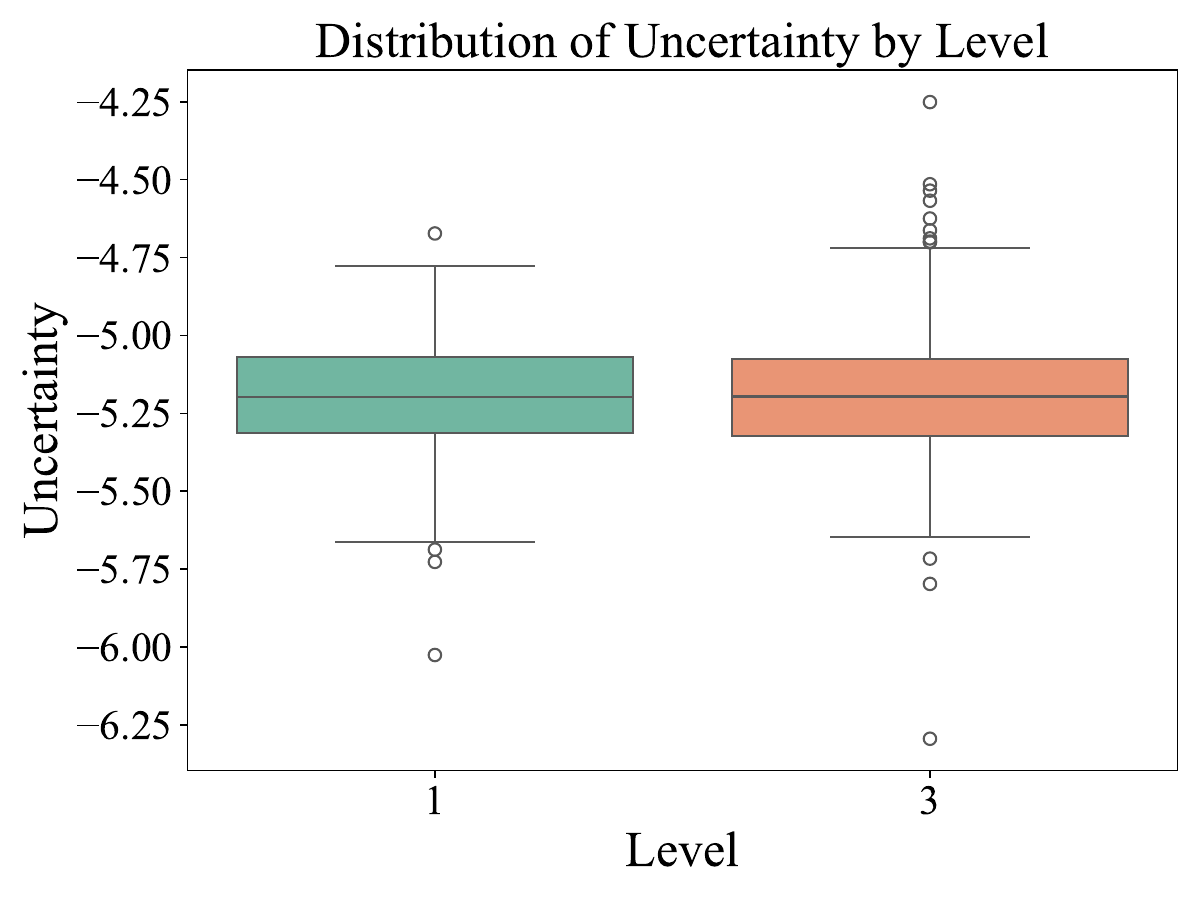} 
    \caption{Projected Bounding Box ($y^t$)}
  \end{subfigure}
  \begin{subfigure}{0.3\linewidth}
    \centering
    \includegraphics[width=\linewidth]{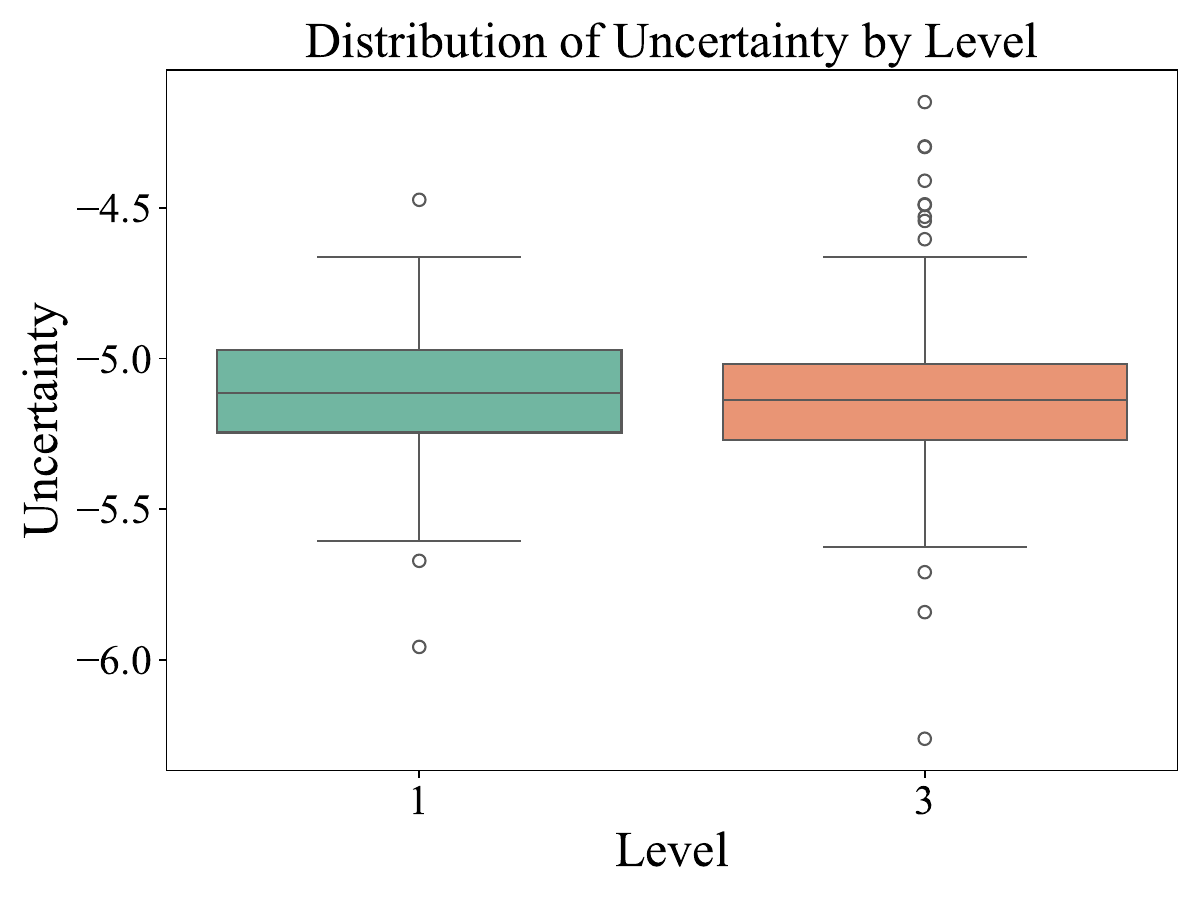} 
    \caption{Projected Bounding Box ($y^b$)}
  \end{subfigure}
  \caption{\textbf{Distribution of predicted uncertainty across difficulty levels on the KITTI validation set.} The x-axis represents the difficulty levels defined by the KITTI benchmark, where level 1 corresponds to \textit{Easy} and level 3 corresponds to \textit{Hard}. The y-axis shows the distributios and mean values of uncertainties predicted by the depth and projected bounding box detection heads. }
  \label{fig:diff_unc}
\end{figure*}

To empirically validate that the predicted uncertainty reflects the instance-level difficulty, we analyze the uncertainty distribution across difficulty levels provided by the KITTI benchmark, which manually annotates each object as \emph{Easy} (Level 1), \emph{Moderate} (Level 2), or \emph{Hard} (Level 3) based on occlusion, truncation, and distance. As shown in Fig.~\ref{fig:diff_unc}, the uncertainty for depth and horizontal box coordinates ($x^l$, $x^r$) increases with difficulty level, confirming that our predicted uncertainty reflects instance-level detection difficulty. In contrast, vertical coordinates ($y^t$, $y^b$) exhibit relatively stable uncertainty across difficulty levels.

This asymmetry stems from the inherent nature of monocular 3D detection: horizontal localization is more affected by occlusion and truncation, whereas vertical positioning is less influenced by these difficulty factors due to by stable geometric contexts such as the ground plane and camera height. Nonetheless, vertical cues remain essential to defining the overall 3D box geometry and provide complementary information, particularly in scenes with sloped terrain or uneven ground.

To further validate this, Tab.~\ref{Table:horizon_vertical_depth} presents a quantitative analysis that compares the impact of applying DAP to horizontal and/or vertical coordinates of the \emph{reparameterized} projected bounding box. Note that these results extend the analysis of Tab.~4 in the main paper. The results show that perturbing horizontal coordinates leads to greater performance gains, consistent with their higher sensitivity to instance-level difficulty. Nevertheless, incorporating perturbations along the vertical axis further yields meaningful improvements, supporting our choice to model uncertainty across both axes. This comprehensive treatment enables more robust perturbation and reconstruction under diverse scene geometries.

\begin{table}[htbp]
\caption{
\textbf{Effect of Different Geometric Supervision Components on Validation Performance.} The last row corresponds to our method using all geometric components.}
\vspace{-8pt}
	\centering
	\scalebox{0.66}{
\begin{tabular}{c|c|c|c| clcc|clcc}
\toprule
\multirow{2}{*}{Class}
% &\multicolumn{3}{c|}{Setting} 
&\multirow{2}{*}{Depth}
& \multicolumn{2}{c}{Bbox (Proj.)} 

% &\multirow{2}{*}{Perturb.}
% &\multirow{2}{*}{$L_{recon}^{bbox}$}
& \multicolumn{4}{c|}{Val, $AP_{BEV|R40}$} 
& \multicolumn{4}{c}{Val, $AP_{3D|R40}$} 
\\ 
 
& & Horizon & Vertical
& \multicolumn{2}{c}{Easy} & Mod. & Hard 
& \multicolumn{2}{c}{Easy} & Mod. & Hard 
\\ \midrule

O & $\times$ & $\times$ & O
& \multicolumn{2}{c}{40.21} & 29.36 & 25.71
& \multicolumn{2}{c}{31.17} & 22.42 & 19.14 \\

O & $\times$ & O & $\times$
& \multicolumn{2}{c}{39.27} & 29.35 & 25.76
& \multicolumn{2}{c}{32.08} & 23.65 & 20.39 \\

O & $\times$ & O & O
& \multicolumn{2}{c}{40.06} & 29.65 & 27.03
& \multicolumn{2}{c}{31.52} & 23.74 & 20.61 \\

O & O & $\times$ & O
& \multicolumn{2}{c}{\underline{40.92}} & 29.92 & 26.37
& \multicolumn{2}{c}{30.34} & 23.00 & 20.04 \\

O & O & O & $\times$
& \multicolumn{2}{c}{40.82} & \underline{29.95} & \underline{27.13}
& \multicolumn{2}{c}{\underline{32.35}} & \underline{24.05} & \underline{20.78} \\

O & O & O & O
& \multicolumn{2}{c}{\textbf{41.68}} & \textbf{30.53} & \textbf{27.76}
& \multicolumn{2}{c}{\textbf{34.89}} & \textbf{25.19} & \textbf{21.78} \\

\bottomrule
\end{tabular}
}

\label{Table:horizon_vertical_depth}
\end{table}

\subsection{Geometric Constraints on Projected Bounding Box Perturbation}

In this section, we explicitly demonstrate that the \emph{reparameterized} Projected Bounding Box Perturbation defined in Eq.~(5) of the main paper consistently satisfies valid geometric constraints. Specifically, we prove that the perturbed coordinates $\tilde{b}^{2D} = (\tilde{x}^l, \tilde{y}^t, \tilde{x}^r, \tilde{y}^b)$ always preserve the correct positional order with respect to the projected bounding box center $(x^{proj}, y^{proj})$, such that:
\begin{equation}
     \tilde{x}^l < x^{proj} < \tilde{x}^r, \quad  \tilde{y}^t < y^{proj} < \tilde{y}^b
\end{equation}

Before perturbaton, the \emph{reparameterized} projected bounding box $b^{2D}=(x^l, y^t, x^r, y^b)$ satisfies the following geometric constraints:
\begin{equation}
\begin{aligned}
    0 &\leq x^l = x^{proj}-o^l < x^{proj} < x^{r}=x^{proj}+o^r \leq 1, \\
    0 &\leq y^t = y^{proj}-o^t < y^{proj} < y^{b}=y^{proj}+o^b \leq 1.
\end{aligned}
\end{equation}

The perturbation scale $\Delta^v$ injected into each coordinate is defined as:

\begin{equation}
\begin{aligned}
    \Delta^{v} = o^v \cdot \hat{c}^v \cdot s^v \cdot \gamma^b, \quad v \in \{l,t,r,b\} 
\end{aligned}
\end{equation}
where $s^{v} \sim U\{-1, 1\}$, $0 \leq \hat{c}^v \leq 1$, and $0 < \gamma^b < 1$. Thus, the absolute value of the perturbation is upper bounded by:
\begin{equation}
\begin{aligned}
    \vert\Delta^{v}\vert = o^v \cdot \hat{c}^v \cdot \gamma^b <  o^v
\end{aligned}
\end{equation}

\noindent Here, $o^v$ for $v\in\{l,t,r,b\}$ denotes the distance from the projected center coordinates $(x^{proj}, y^{proj})$ to each side of the bounding box. This inequality indicates that the perturbation magnitude $\vert\Delta^{v}\vert$ is always less than the corresponding boundary distance $o^v$, ensuring that the perturbed coordinates remain within valid geometric limits.

Specifically, the perturbed coordinates satisfy the following conditions. For the horizontal coordinates $x^l$, the perturbation in the positive direction yields:
\begin{equation}
\begin{aligned}
    \tilde{x}^l = x^l + \Delta^l < x^l + o^l = x^{proj}
\end{aligned}
\end{equation}

Conversely, perturbing negatively gives:
\begin{equation}
\begin{aligned}
    \tilde{x}^l = x^l + \Delta^l > x^l - o^l = x^{proj} - 2o^l
\end{aligned}
\end{equation}
Here, $x^{proj} - 2o^l$ could potentially yield a negative value; however, the \texttt{CLIP} in Eq. (5) of the main paper operation enforces a lower bound of 0, thereby guaranteeing that $0 \leq \tilde{x}^l < x^{proj}$ always holds.

The same logic applies to the right boundary coordinate $x^r$. For the maximal negative perturbation:
\begin{equation}
\begin{aligned}
    \tilde{x}^r = x^r + \Delta^r > x^r - o^r = x^{proj}
\end{aligned}
\end{equation}

\noindent and for the maximal positive perturbation:
\begin{equation}
\begin{aligned}
    \tilde{x}^r = x^r + \Delta^r  < x^r + o^r = x^{proj} + 2o^r
\end{aligned}
\end{equation}

$x^{proj} + 2o^r$ may also exceed 1, the \texttt{CLIP} operation restricts the upper bound to 1, ensuring $x^{proj} < \tilde{x}^r \leq 1$.

Similarly, the top and bottom boundary coordinates, $\tilde{y}^t$ and $\tilde{y}^b$, are guaranteed to satisfy the geometric constraints $0 \leq \tilde{y}^t < y^{proj}$ and $y^{proj} < \tilde{y}^b \leq 1$, respectively.

Consequently, the perturbed bounding box coordinates are guaranteed to satisfy the following geometric constraints:
\begin{equation}
\begin{aligned}
    0 &\leq \tilde{x}^l < x^{proj} < \tilde{x}^{r} \leq 1  
    \rightarrow 0 \leq \tilde{x}^l < \tilde{x}^{r} \leq 1\\
    0 &\leq \tilde{y}^t < y^{proj} < \tilde{y}^{b} \leq 1 
    \rightarrow 0 \leq \tilde{y}^t < \tilde{y}^{b} \leq 1
\end{aligned}
\end{equation}

These inequalities ensure that the perturbed box remains a valid bounding box with properly ordered and normalized coordinates.

\subsection{Implementation Details} 

\paragraph{Implementation Details of Perturbed Label Query}
Following the approaches introduced by DN-DETR~\cite{dndetr} and DINO~\cite{zhang2022dino}, we construct multiple perturbation groups when generating perturbed label queries. Within each perturbation group, both positive and negative queries are created. 

Specifically, each query is perturbed by randomly sampling a perturbation sign $s^v \sim U\{-1,1\}$, and multiplying it by a group agnostic perturbation scale $o^v\cdot\hat{c}^v\cdot \gamma^b$, ensuring each query group to have diverse perturbations signs while sharing the same perturbation scale.
Consequently, the queries within each perturbation group reflect varied perturbations relative to the original labels. The effect of the number of perturbation groups on model performance is analyzed in Tab.~\ref{Table:Group_Number}.

Additionally, following DINO~\cite{zhang2022dino}, we generate negative query groups corresponding to each positive perturbation group.
Specifically, for negative queries, we increase the perturbation magnitude by adding 1 to the perturbation scale 
 of the corresponding positive queries, preserving the original perturbation direction defined by the sign $s^v$.

During training, these negative queries are explicitly assigned to the "no object" class in the class reconstruction loss, effectively encouraging the model to distinguish foreground objects from background regions more clearly.

\paragraph{Implementation Differences between MonoDGP and MonoDETR}
In the main paper, we demonstrated that our proposed method can be effectively integrated with DETR-based detectors such as MonoDETR~\cite{monodetr} and MonoDGP~\cite{monodgp}. Although both methods share the same DETR architecture, they differ in how the final depth prediction is derived from the outputs of their respective depth heads. These differences lead to distinct strategies for constructing the label queries for the depth component. Below, we explicitly outline these implementation differences:

\begin{itemize}
    \item \textbf{Depth Predictoin in MonoDETR:} MonoDETR computes the final predicted depth $d_{pred}$ as an average of three components: the direct output from the depth head ($d_{reg}$), the geometric depth derived from perspective projection formula ($d_{geo}$), and a foreground depth map estimation ($d_{map}$). Since the depth head directly predicts the absolute regressed depth value, we use the ground-truth depth as the depth component when generating label queries.

    \item \textbf{Depth Prediction in MonoDGP:} MonoDGP treats the output of the depth head as a residual error term ($d_{err}$) rather than an absolute depth value. The final predicted depth $d_{pred}$ is calculated by adding this error term to the geometric depth $d_{geo}$:
    \begin{equation}
     d_{pred} = d_{gro} + d_{err}   
    \end{equation}
    where $d_{err}$ indicates the discrepency between the actual ground truth depth and geometric depth. Given that the MonoDGP depth head outputs this error term, we construct the label query for the depth head by using the residual error computed from the ground truth depth and the geometric depth ground truth.

\end{itemize}

\subsection{Additional Ablation Studies} 

% EMA scale factor
\paragraph{EMA cofficients for difficulty score 
min/max normalization.}

\noindent In Eq. (4) of the main paper, we apply EMA to update the minimum and maximum values used for normalizing the difficulty scores, effectively capturing the global distribution of data. Tab.~\ref{Table:EMA_momentum} shows the ablation results for different EMA momentum coefficients. The baseline corresponds to batch-wise min/max normalization without EMA, resulting in performance degradation due to its inability to reflect the global distribution. The best performance was achieved with an momentum coefficient of 0.8, which we adopt as the default setting.

\begin{table}[!ht]
\caption{\textbf{Ablation study of the EMA momentum coefficient.} The baseline corresponds to the case without EMA.}
\vspace{-8pt}
	\centering
	\small
	\scalebox{0.8}{
\begin{tabular}{c|clcc|clcc}
\toprule
\multicolumn{1}{l|}{\multirow{2}{*}{Momentum}} 
& \multicolumn{4}{c|}{Val, $AP_{BEV|R40}$} 
& \multicolumn{4}{c}{Val, $AP_{3D|R40}$} 
\\ 
& \multicolumn{2}{c}{Easy} & Mod. & Hard 
& \multicolumn{2}{c}{Easy} & Mod. & Hard 
\\ \midrule

Baseline
& \multicolumn{2}{c}{41.44} & 31.33 & 27.61 
& \multicolumn{2}{c}{32.22} & 24.32 & 20.94 \\
\midrule
$\beta=0.7$
& \multicolumn{2}{c}{41.41} & 31.37 & 27.84
& \multicolumn{2}{c}{33.24} & 24.23 & 21.78 \\
$\beta=0.8$ 
& \multicolumn{2}{c}{\textbf{41.68}} & \textbf{30.53} & \textbf{27.76}
& \multicolumn{2}{c}{\textbf{34.89}} & \textbf{25.19} & \textbf{21.78} \\
$\beta=0.9$ 
& \multicolumn{2}{c}{41.71} & 30.55 & 28.06
& \multicolumn{2}{c}{33.2} & 24.29 & 21.16 \\
\bottomrule

\end{tabular}
}
\label{Table:EMA_momentum}
\end{table}
\vspace{+8pt}

\paragraph{Stage 1 Input: Ground-Truth vs. Uniformly-Corrupted Label Queries.}

In Tab.~\ref{Table:Label_query}, we compare the detection performance using two types of label queries for Stage 1 of Fig.~3 in the main paper: ground-truth labels and uniformly-corrupted labels. These label queries serve as decoder inputs for the Difficulty-Aware Perturbation (DAP) module in Stage 1. The first row (\textit{Label (Ours)}) corresponds to using noise-free ground-truth labels to estimate instance-level detection uncertainty. The second row (\textit{Uniform Noised Label}) denotes queries constructed by injecting uniform noise into the labels. Experimental results show that using clean labels yields consistently better performance than the noisy counterpart. 
This is becuase ground-truth labels inherently encode precise object-level geometry and supervision fidelity, thereby providing a more stable and reliable signal for accurate uncertainty estimation.
Additionally, our approach avoids the need for extensive hyperparameter tuning (e.g., uniform noise scale selection), resulting in a more stable and efficient training process.

\begin{table}[ht!]
\caption{\textbf{Impact of Noise Injection on Label Queries.}}

\vspace{-8pt}
	\centering
	\small
	\scalebox{0.8}{
\begin{tabular}{l|clcc|clcc}
\toprule
\multicolumn{1}{l|}{\multirow{2}{*}{Method}} 
& \multicolumn{4}{c|}{Val, $AP_{BEV|R40}$} 
& \multicolumn{4}{c}{Val, $AP_{3D|R40}$} 
\\ 
& \multicolumn{2}{c}{Easy} & Mod. & Hard 
& \multicolumn{2}{c}{Easy} & Mod. & Hard 
\\ \midrule

Label (Ours)
& \multicolumn{2}{c}{41.68} & 30.53 & 27.76
& \multicolumn{2}{c}{34.89} & 25.19 & 21.78 \\
\midrule
Uniform Noised Label
& \multicolumn{2}{c}{39.17} & 29.57 & 25.96
& \multicolumn{2}{c}{31.68} & 23.85 & 20.55 \\
\bottomrule

\end{tabular}
}
\label{Table:Label_query}
\end{table}
\vspace{+8pt}

\paragraph{Ablation Study of Perturbation Scaling Factors} % depth, bbox scale

We analyze the effects of two perturbation scale factors used during the Difficulty-Aware Perturbation (DAP) process in Stage 1: $\gamma^b$ , which controls the perturbation magnitude of the projected bounding box, and $\gamma^d$, which determine the perturbation magnitude applied to the depth label. Tab.~\ref{Table:Bbox_Scale_Factor} and Tab.~\ref{Table:Depth_Scale_Factor} show the effects of these scale factors. The best performance is achieved when $\gamma^d$ and $\gamma^b$ are set to 0.8 and 0.4, respectively. This suggests that reconstructing perturbed bounding box coordinates is inherently more challenging than reconstructing perturbed depth values. 

Moreover, our depth perturbation strategy employs the depth error—defined as the discrepancy between geometric depth and actual ground-truth depth—as the target depth label (detailed in Supplementary Section D). Perturbing this relative error, rather than the absolute depth, inherently provides greater numerical stability. Specifically, depth errors typically exhibit small numerical magnitudes (approximately within the range of -2 to 2 in practice), thus allowing appropriate perturbation scales without significantly distorting the original error distribution. As a result, applying a relatively larger perturbation scale factor to depth labels enhances training effectiveness and contributes positively to overall detection performance and model robustness.

\begin{table}[ht!]
\caption{\textbf{Ablation study of the Bounding box Scale Factor $\gamma^b$.}}
\vspace{-8pt}

	\centering
	\small
	\scalebox{0.8}{
\begin{tabular}{c|clcc|clcc}
\toprule
\multicolumn{1}{l|}{\multirow{2}{*}{$\gamma^b$}} 
& \multicolumn{4}{c|}{Val, $AP_{BEV|R40}$} 
& \multicolumn{4}{c}{Val, $AP_{3D|R40}$} 
\\ 
& \multicolumn{2}{c}{Easy} & Mod. & Hard 
& \multicolumn{2}{c}{Easy} & Mod. & Hard 
\\ \midrule

0.2
& \multicolumn{2}{c}{39.32} & 29.94 & 27.4
& \multicolumn{2}{c}{32.02} & 24.22 & 21.81 \\

\textbf{0.4}
& \multicolumn{2}{c}{\textbf{41.68}} & \textbf{30.53} & \textbf{27.76}
& \multicolumn{2}{c}{\textbf{34.89}} & \textbf{25.19} & \textbf{21.78} \\

0.6 
& \multicolumn{2}{c}{42.38} & 31.88 & 28.23
& \multicolumn{2}{c}{33.28} & 24.44 & 21.17 \\

0.8 
& \multicolumn{2}{c}{40.24} & 29.77 & 26.17
& \multicolumn{2}{c}{32} & 23.85 & 20.59 \\
\bottomrule

\end{tabular}
}
\label{Table:Bbox_Scale_Factor}
\end{table}

\begin{table}[ht!]
\caption{\textbf{Ablation study of the Depth Scale Factor $\gamma^d$.}}
\vspace{-8pt}
	\centering
	\small
	\scalebox{0.8}{
\begin{tabular}{c|clcc|clcc}
\toprule
\multicolumn{1}{l|}{\multirow{2}{*}{$\gamma^d$}} 
& \multicolumn{4}{c|}{Val, $AP_{BEV|R40}$} 
& \multicolumn{4}{c}{Val, $AP_{3D|R40}$} 
\\ 
& \multicolumn{2}{c}{Easy} & Mod. & Hard 
& \multicolumn{2}{c}{Easy} & Mod. & Hard 
\\ \midrule

1.0
& \multicolumn{2}{c}{40.68} & 30.34 & 27.85 
& \multicolumn{2}{c}{32} & 24.29 & 21.21 \\

\textbf{0.8}
& \multicolumn{2}{c}{\textbf{41.68}} & \textbf{30.53} & \textbf{27.76}
& \multicolumn{2}{c}{\textbf{34.89}} & \textbf{25.19} & \textbf{21.78} \\

0.6 
& \multicolumn{2}{c}{42.59} & 30.59 & 26.59
& \multicolumn{2}{c}{34.03} & 24.68 & 21.15 \\

0.4 
& \multicolumn{2}{c}{43.39} & 30.56 & 26.51
& \multicolumn{2}{c}{33.41} & 24 & 20.54 \\
\bottomrule

\end{tabular}
}
\label{Table:Depth_Scale_Factor}
\end{table}

  \paragraph{Ablation Study of on the Number of Perturbation Groups}

In Tab.~\ref{Table:Group_Number}, we analyze the influence of the number of perturbation groups used to construct perturbed label queries. 
Here, the number of perturbation groups refers to $M$, with a fixed number of original label queries $K$. Specifically, when the group number $N=1$,  the total number of perturbed label queries is $2 \times K$, consisting of one positive and one negative set of queries. In general, the total number of perturbed label queries is calculated as $2 \times N \times K$, proportional to the selected perturbation groups $N$.
Each perturbation group is constructed based on the randomness of perturbation sign $s^v$, which enables the diverse perturbation patterns. Experimental results show that using 7 perturbation groups yields the best performance. This suggests that the randomness of perturbation signs is effective in producing sufficiently diverse perturbed label queries for robust learning.

\begin{table}[!ht]
\caption{\textbf{The Number of Perturbed Label Query Groups.}}
\vspace{-8pt}
	\centering
	\small
	\scalebox{0.8}{
\begin{tabular}{c|clcc|clcc}
\toprule
\multicolumn{1}{l|}{{\# of Perturbed}} 
& \multicolumn{4}{c|}{Val, $AP_{BEV|R40}$} 
& \multicolumn{4}{c}{Val, $AP_{3D|R40}$} 
\\ 
\multicolumn{1}{l|}{{Label Query Groups}} 
& \multicolumn{2}{c}{Easy} & Mod. & Hard 
& \multicolumn{2}{c}{Easy} & Mod. & Hard 
\\ \midrule

1
& \multicolumn{2}{c}{40.63} & 29.66 & 26.05 
& \multicolumn{2}{c}{32.58} & 23.75 & 20.56 \\

3
& \multicolumn{2}{c}{41.27} & 30.13 & 27.45
& \multicolumn{2}{c}{32.88} & 24.37 & 21.09 \\

5
& \multicolumn{2}{c}{41.08} & 29.83 & 26.40
& \multicolumn{2}{c}{33.21} & 24.02 & 20.94 \\

\textbf{7}
& \multicolumn{2}{c}{\textbf{41.68}} & \textbf{30.53} & \textbf{27.76}
& \multicolumn{2}{c}{\textbf{34.89}} & \textbf{25.19} & \textbf{21.78} \\

8
& \multicolumn{2}{c}{42.16} & 31.17 & 27.59
& \multicolumn{2}{c}{31.91} & 23.93 & 20.76 \\

\bottomrule
\end{tabular}
}
\label{Table:Group_Number}
\end{table}

\begin{figure*}[!th]
  \centering
    \includegraphics[width=0.99\linewidth]{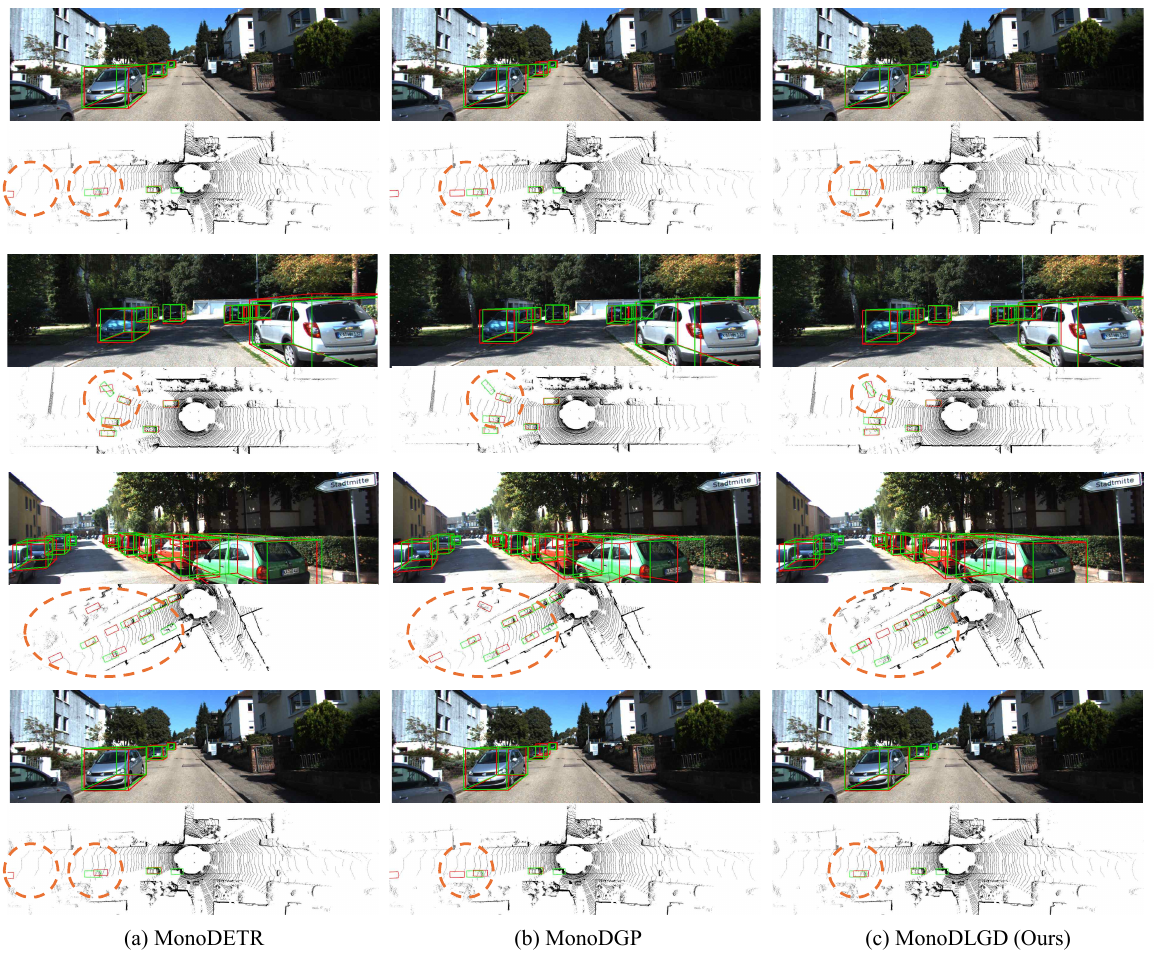}   
  \caption{\textbf{Quality Analysis on KITTI val dataset.} (a) MonoDETR~\cite{monodetr} (b) MonoDGP~\cite{monodgp} (c) MonoDLGD (Ours). For each image, \textcolor{green}{Green} box represents ground truth box and \textcolor{red}{Red} box represents predicted value.}
  \label{fig:BEV_Vis}
\end{figure*}

\subsection{Training Complexity}  
% Memory Usage 부분 추가
In Tab.~\ref{Table:train_complexity}, we analyze the training time and memory usage overhead introduced by our MonoDLGD framework. When applied to the baseline model, MonoDGP~\cite{monodgp}, the training time increases by approximately 17.4 \%, and memory usage increased by approximately 5.99GB.
These increases primarily stem from the additional computational steps introduced by Difficulty-Aware Perturbation (DAP) module, including difficulty score computation, perturbation injection, and the subsequent label reconstruction processes. Additionally, memory usage grows due to the utilization of multiple perturbed label query groups, in this case, seven groups, as discussed in Section E.4.

However, as demonstrated by the computational cost analysis in Tab.~2 of the main paper, these operations are only required during training. The inference process remains unaffected by DAP, resulting in negligible inference-time overhead. Therefore, the proposed MonoDLGD framework delivers substantial accuracy improvements without sacrificing inference efficiency.

\begin{table}[!htb]
\caption{
\textbf{Training Complexity.} 
}
\vspace{-8pt}
	\centering
	\scalebox{0.75}{
\begin{tabular}{l|clcc|c|c}
\toprule
\multicolumn{1}{l|}{\multirow{2}{*}{Method}}  
& \multicolumn{4}{c|}{Val, $AP_{3D|R40}$} 

& Training time & Memory
\\ 
& \multicolumn{2}{c}{Easy} & Mod. & Hard & (min/epoch)$\downarrow$ & Usage (GB)$\downarrow$
\\ \midrule

MonoDGP

& \multicolumn{2}{c}{30.76} & 22.34 & 19.02 
& 4.55 & 12.35\\
MonoDGP+Ours
& \multicolumn{2}{c}{34.89} & 25.19 & 21.78
& 5.51 & 18.34 \\

\bottomrule
\end{tabular}
}
\label{Table:train_complexity}
\end{table}

\subsection{More Results} 

\noindent\textbf{Experiments on Other Categories}

Tab.~\ref{tab:val_other} presents additional experimental results on other object categories (Pedestrian and Cyclist) from the KITTI dataset. 
Compared to MonoDGP, MonoDLGD consistently improves performance on the Cyclist category across all difficulty levels, achieving notalbe AP improvements of 4.01 (Easy), 1.86 (Moderate), and 1.28 (Hard). 
However, for the Pedestrian category, MonoDLGD provide 0.02 AP improvement at the Moderate level, while MonoDGP achieves slightly better performance in Easy and Hard levels.
% test 결과 추가.
Tab.~\ref{Table:Cylist} presents the performance on the KITTI test set. MonoDLGD consistently achieves competitive results across all difficulty levels for the Cyclist category. For the Pedestrian category, MonoDLGD demonstrates the best performance among compared methods, but the performance improvements are relatively modest compared to those observed for Cyclist.
This discrepancy in performance between categories can be explained by the uncertainty analysis presented in Fig.~\ref{fig:diff_unc}. Specifically, the difficulty levels defined by the KITTI benchmark primarily reflect challenges related to horizontal localization, including occlusion and truncation. Consequently, our approach, which explicitly models horizontal localization uncertainty, excels on object categories such as Car and Cyclist that exhibit larger horizontal dimensions. Conversely, objects like Pedestrians, which typically possess a vertical-elongated structure, experience relatively stable vertical localization uncertainty.
In summary, the proposed MonoDLGD effectively enhances performance for object categories heavily influenced by horizontal localization uncertainty. However, its impact is relatively limited for vertically elongated or upright objects. This observation suggests future research directions toward category-specific perturbation strategies that account for the distinctive geometric properties of different object types.

\begin{table}[!thb]
\caption{Ablation study of the pedestrian and cyclist categories on the KITTI val set.}
\centering
\resizebox{1.0\columnwidth}{!}{
\begin{tabular}{l|cccccc}
\toprule
\multirow{3}{*}{Methods} & \multicolumn{6}{c}{Val, IoU=0.5, $AP_{3D|R40}$} \\ \cline{2-7} 
 & \multicolumn{3}{c|}{Pedestrian} & \multicolumn{3}{c}{Cyclist} \\ \cline{2-7} 
 & Easy & Mod. & \multicolumn{1}{c|}{Hard} & Easy & Mod. & Hard \\ \midrule
MonoDGP \cite{monodgp} & \textbf{13.77} & 10.06 & \multicolumn{1}{c|}{\textbf{7.96}} & 12.21 & 6.61 & 5.95 \\
\midrule
MonoDLGD (Ours)& 13.04 & \textbf{10.08} & \multicolumn{1}{c|}{7.56} & \textbf{16.22} & \textbf{8.47} & \textbf{7.23} \\

\bottomrule

\end{tabular}
}
\label{tab:val_other}
\end{table}

\begin{table}[h!]
\caption{Comparisons of the pedestrian and cyclist categories on the KITTI test set. }
\vspace{-8pt}
\centering
\resizebox{1.0\columnwidth}{!}{%
\begin{tabular}{l|c|cccccc}
\hline
\multirow{3}{*}{Methods} & \multirow{3}{*}{\begin{tabular}[c]{@{}c@{}}Extra \\ data\end{tabular}} & \multicolumn{6}{c}{Test, IoU=0.5, $AP_{3D|R40}$ } \\ \cline{3-8} 
 &  & \multicolumn{3}{c|}{Pedestrian} & \multicolumn{3}{c}{Cyclist} \\ \cline{3-8} 
 &  & Easy & Mod. & \multicolumn{1}{c|}{Hard} & Easy & Mod. & Hard \\ \hline
CaDDN~\cite{caddn} & \multirow{2}{*}{LiDAR} & 12.87 & 8.14 & \multicolumn{1}{c|}{6.76} & {7.00} & 3.41 & {3.30} \\
OccupancyM3D~\cite{occupancym3d} &  & 14.68 & 9.15 & \multicolumn{1}{c|}{7.80} & \textbf{7.37} & {3.56} & 2.84 \\ \hline
MonoPGC~\cite{monopgc} & Depth & 14.16 & {9.67} & \multicolumn{1}{c|}{{8.26}} & 5.88 & 3.30 & {2.85} \\ \hline
GUPNet~\cite{gupnet} & \multirow{4}{*}{None} & {14.72} & 9.53 & \multicolumn{1}{c|}{7.87} & 4.18 & 2.65 & 2.09 \\
MonoCon~\cite{monocon} &  & 13.10 & 8.41 & \multicolumn{1}{c|}{6.94} & 2.80 & 1.92 & 1.55 \\
DEVIANT~\cite{deviant} &  & 13.43 & 8.65 & \multicolumn{1}{c|}{7.69} & 5.05 & 3.13 & 2.59 \\
MonoDDE~\cite{monodde} &  & 11.13 & 7.32 & \multicolumn{1}{c|}{6.67} & 5.94 & 3.78 & 3.33 \\
MonoDETR~\cite{monodetr} &  & 12.65 & 7.19 & \multicolumn{1}{c|}{6.72} & 5.12 & 2.74 & 2.02 \\
MonoDGP~\cite{monodgp} &  & 15.04 & 9.89 & \multicolumn{1}{c|}{8.38} & 5.28 & 2.82 & 2.65 \\ \hline
MonoDLGD (Ours) & None & \textbf{15.99} & \textbf{10.44} & \multicolumn{1}{c|}{\textbf{8.82}} & \textbf{6.62} & \textbf{4.39} & \textbf{3.35} \\ \hline
\end{tabular}
}
\label{Table:Cylist}
\end{table}
\vspace{+8pt}

\subsection{Qualitative Results} 

To intuitively evalutate the effectiveness of our model under vairous difficulty conditions, we visualize 3D detection results and bird's-eye view (BEV) representations on the KITTI validation set. In Fig.~\ref{fig:BEV_Vis}, we compare the detection results of previous DETR-based baselines; (a) MonoDETR, (b) MonoDGP, and (c) MonoDLGD (Ours). MonoDLGD achieves more accurate and consistent detection performance, even in challenging environments involving distant or occluded objects.

%\bibliography{aaai2026}

\end{document}